\documentclass[sigconf]{acmart}

\usepackage{amsmath}
\usepackage[utf8]{inputenc}
\usepackage{pgfplots}
\usepgfplotslibrary{patchplots,colormaps}
\DeclareUnicodeCharacter{2212}{−}
\usepgfplotslibrary{groupplots,dateplot}
\usetikzlibrary{patterns,shapes.arrows,positioning,shapes.geometric,fit, trees}
\pgfplotsset{compat=newest}
\usepackage{adjustbox}
\usepackage{csquotes}

\usepackage{appendix}
\usepackage{subcaption}
\usepackage{xcolor}
\usepackage{listings}
\usepackage{url}
\usepackage{graphicx}
\usepackage{hyperref}
\hypersetup{colorlinks=true,linkcolor=blue,urlcolor=blue, linkbordercolor=red}

\usepackage{cleveref}
\usepackage{tabularx}
\usepackage{drawstack}

\crefdefaultlabelformat{\color{red}(#2#1#3)}
\newcommand{\code}[1]{\texttt{#1}}
\newcommand{\link}[1]{\color{blue}#1}

\usepackage[cache=false, draft]{minted}
\newmintedfile{c}{
linenos=TRUE,
fontsize=\small,
numbersep=4pt,
}

\definecolor{cudnn}{rgb}{       0.88627,0.29019,0.2}
\definecolor{libtorch}{rgb}{    0.20392,0.54117,0.74117}
\definecolor{pytorch}{rgb}{     0.59607,0.55686,0.83529}
\definecolor{torchscript}{rgb}{ 0.99607,0.55686,0.83529}

\makeatletter
\newcommand*\bigcdot{\mathpalette\bigcdot@{.5}}
\newcommand*\bigcdot@[2]{\mathbin{\vcenter{\hbox{\scalebox{#2}{$\m@th#1\bullet$}}}}}
\makeatother

\begin{document}

\title{\textsc{Comparing the costs of abstraction for DL frameworks}}

\author{Maksim Levental}
\authornote{Both authors contributed equally to this work.}
\email{mlevental@uchicago.edu}
\affiliation{%
  \institution{University of Chicago}
}
\author{Elena Orlova}
\email{eorlova@uchicago.edu}
\affiliation{%
  \institution{University of Chicago}
}

\renewcommand{\shortauthors}{Levental and Orlova}

\begin{abstract}
  High level abstractions for implementing, training, and testing Deep Learning (DL) models abound.
  Such frameworks function primarily by abstracting away the implementation details of arbitrary neural architectures, thereby enabling researchers and engineers to focus on design.
  In principle, such frameworks could be ``zero-cost abstractions'';
  in practice, they incur translation and indirection overheads.
  We study at which points exactly in the engineering life-cycle of a DL model the highest costs are paid and whether they can be mitigated.
  We train, test, and evaluate a representative DL model using PyTorch, LibTorch, TorchScript, and cuDNN on representative datasets, comparing accuracy, execution time and memory efficiency.
\end{abstract}

\maketitle

\section{Introduction}\label{sec:introduction}
Deep Learning (DL) frameworks represent neural network models as dataflow and computation graphs (where nodes correspond to functional units and edges correspond to composition).
In recent years, there has been a proliferation of DL frameworks~\cite{paszke2019pytorch,abadi2016tensorflow,chen2015mxnet,cntk} implemented as domain-specific languages (DSLs) embedded in ``high-level'' languages%
\footnote{For the purposes of this article, we take ``high-level'' to mean garbage collected and agnostic with respect to hardware \textit{from the perspective of the user}.} such as Python, Java, and C\#.
These DSLs serve as \textit{abstractions} that aim to map the DL graphs to hardware pipelines.
That is to say, they hide (or \textit{encapsulate}) details of DL models that are judged to be either irrelevant or too onerous to consider (see~\cref{subsec:abstraction} for a more comprehensive discussion on abstraction in computer science).
By virtue of these design decisions the frameworks trade-off ease-of-use for execution performance;
quoting the architects of PyTorch:
\begin{displayquote}[\cite{paszke2019pytorch}]
  To be useful, PyTorch needs to deliver compelling performance, although not at the expense of simplicity and ease of use.
  Trading 10\% of speed for a significantly simpler to use model is acceptable; 100\% is not.
\end{displayquote}

Trading off ergonomics for performance is manifestly reasonable%
\footnote{\textcquote{knuth}{The real problem is that programmers have spent far too much time worrying about efficiency in the wrong places and at the wrong times; premature optimization is the root of all evil (or at least most of it) in programming.}}%
, especially during the early phases of the DL engineering/research process (i.e.\ during the hypothesis generation and experimentation phases).
Ultimately one needs to put the DL model into production.
It is at this phase of the DL engineering process that every percentage point of execution performance becomes critical.
Alternatively, there are many areas of academic DL where the research community strives to incrementally improve performance~\cite{abdelhamed2020ntire,hall2020probability,ILSVRC15}.
For example, in the area of super-resolution a high-priority goal is to be able to ``super-resolve'' in real-time~\cite{7780576}.
Similarly, in natural language processing, where enormous language models are becoming the norm~\cite{brown2020language}, memory efficiency of DL models is of the utmost concern.
In such instances it is natural to wonder whether ease-of-use trade-offs that sacrifice execution performance, or memory efficiency, are worthwhile and whether their costs can be mitigated.

Thus, our intent here is to investigate the costs of some of the abstractions employed by framework developers.
In particular we focus on the PyTorch ecosystem (chosen for its popularity amongst academic researchers) deployed to Graphics Processing Units (GPUs).
To that end, we implement a popular and fairly representative%
\footnote{In the sense that the functional units constituting the model are widely used in various other models.}
DL model at four levels of abstraction: conventional PyTorch, LibTorch, cuDNN, and TorchScript.
We argue, in the forthcoming, that these four implementations span considerable breadth in the abstraction spectrum.
Furthermore we train, test, and evaluate each of the implementations on four object detection datasets and tabulate performance and accuracy metrics.

The rest of this article is organized as follows: ~\cref{sec:background} discusses abstraction and quickly reviews the germaine background material on GPUs and DL frameworks, ~\cref{sec:methodology} describes the implementations and our profiling methodology,~\cref{sec:results} presents our results and a comparative discussion thereof, ~\cref{sec:discussion} discusses broad lessons learned, ~\cref{sec:futurework} concludes and proposes future work, and ~\cref{sec:speculation} speculates wildly about the future of DL systems more generally.


\section{Background}\label{sec:background}
\subsection{Abstraction}\label{subsec:abstraction}
What is abstraction?
In fact, there are several closely related notions of abstraction.
First there is the philosophical notion of abstraction;
Locke defines abstraction as follows (bolding ours):
\begin{displayquote}[\cite{Locke1689-LOCAEC-4}]
    The acts of the mind, wherein it exerts its power over simple ideas, are chiefly these three: \textellipsis
    The third is \textbf{separating them from all other ideas that accompany them in their real existence}: this is called abstraction \textellipsis
\end{displayquote}
Then there is mathematical abstraction;
Russell defines abstraction as follows:
\begin{displayquote}[\cite{Russell1937-RUSPOM-7}]
    This principle [of abstraction] asserts that, whenever a relation, of which there are instances, has the two properties of being symmetrical and transitive, then the relation in question is not primitive, but is analyzable into sameness of relation to some other term;
    and that this common relation is such that there is only one term at most to which a given term can be so related, though many terms may be so related to a given term.
\end{displayquote}
Intriguing as these notions of abstraction may be, they are distinctly different from the notion of abstraction in computer science;
in particular with respect to mathematical abstraction (bolding ours):
\begin{displayquote}[\cite{abstraction}]
    \textellipsis the primary product of mathematics is \textit{inference structures}, while the primary product of computer science is \textit{interaction patterns}.
    This is a crucial difference, and it shapes their use of formalism and the kind of abstraction used in the two disciplines.

    \vspace{4pt}

    \textellipsis computer science is distinguished from mathematics in the use of a kind of abstraction that computer scientists call \textit{information hiding}.
    The complexity of behaviour of modern computing devices makes the task of programming them impossible without abstraction tools that hide, but do not neglect, \textbf{details that are essential in a lower-level processing context but inessential in a [particular] software design and programming context}.
\end{displayquote}
This understanding of abstraction is widely agreed upon;
notably Abelson, Sussman, and Sussman in their much revered \textit{Structure and Interpretation of Programs}:
\begin{displayquote}[\cite{abelson1996structure}]
    We are not at that moment concerned with how the procedure computes its result, only with the fact that it computes the square.
    The details of how the square is computed can be suppressed, to be considered at a later time.
    Indeed, as far as the \code{good-enough?} procedure is concerned, \code{square} is not quite a procedure but rather an abstraction of a procedure, a so-called \textit{procedural abstraction}.
    At this level of abstraction, any procedure that computes the square is equally good.
\end{displayquote}

Thus, abstraction is the modulation of concern for details in accordance with the needs of the user and \textit{levels of abstractions} are graded by the degree of the elision (bolding ours):
\begin{displayquote}[\cite{abstraction}]
    To specify nontrivial computational processes in machine language is a practical impossibility for humans, and so programming languages with higher levels of abstraction are necessary.

    \vspace{4pt}

    \textellipsis At a higher level of [abstraction], a \textit{subroutine}, \textit{function}, or \textit{procedure} is an abstraction of a segment of memory that hides the details of how the segment represents a piece of a program that is passed certain parameter values and returns a value as a result.

    \vspace{4pt}

    \textellipsis A \textit{garbage collector} is an abstraction of a special process that collects garbage and makes it once again available to the original program, hiding from that program the details of how this is done.

    \vspace{4pt}

    \textellipsis \textbf{This use of code libraries is an example of \textit{procedural abstraction}}, or the ability to execute code through the calling of named procedures that accept explicitly described parameters and return certain guaranteed results.
    It is an example of abstraction because the details of how the procedure performs its computation are hidden from the procedure's caller;
    since the caller only makes use of the procedure for its results, there is no need for it to know the internals.
\end{displayquote}

Taking \textit{information and concern encapsulation} as our operational definition of abstraction, in what sense shall we measure the costs of the abstractions employed by DL frameworks?
An immediate candidate measure of cost is the asymptotic time (or space) complexity of various operations and data structures that comprise the abstraction.
We claim that, with rare exception\footnote{One result does come to mind: Pippenger~\cite{10.1145/244795.244798} produces a program that runs in O$(n)$ on an impure implementation (i.e. with side-effects) LISP but which runs in $\Theta(n \log n)$ on a pure LISP\@.}, asymptotic complexity is a poorly suited measure of the complexity or cost of abstractions in the sense that we here deal with.
If the abstraction is truly abstract then it bears no resemblance to the realization (recall Locke's definition of abstraction) and if the abstraction is reified then the analysis becomes completely impractical (owing to the numerous components and operations).
Even if such analysis were practicable the result would most likely be uninteresting and inconsequential for actual DL frameworks and their users.
It is well known that the constant factors in the complexity and particularities of hardware systems themselves more closely govern performance than the order terms.
For example, Quicksort, an O$\left(n^2\right)$ sorting routine, outperforms even many $\Theta(n\log n)$ sorting routines because it is more cache efficient~\cite{10.5555/1410219}.

Another way to reason about the cost of abstractions is according to the ``zero-overhead'' principle as articulated by Bjarne Stroustrup:
\begin{displayquote}[\cite{10.1007/978-3-642-28869-2_1}]
    In general, C++ implementations obey the zero-overhead principle: What you don't use, you don't pay for.
    And further: What you do use, you couldn't hand code any better.
\end{displayquote}
Therefore we make the expedient and practical assumption that what is more interesting and valuable to the DL community than asymptotics is, in fact, an empirical study of the resource efficiency of the abstractions;
namely execution time, memory usage, and GPU utilization.

\subsection{GPUs}\label{subsec:gpus}

We briefly review NVIDIA GPUs%
\footnote{A more comprehensive introduction to GPUs themselves and CUDA programming is available in~\cite{10.5555/1891996}.}
in order that the performance criteria we measure in~\cref{sec:methodology} are legible.

\begin{figure}
    \centering
    \begin{subfigure}{\linewidth}
        \centering
        \includegraphics[width=\linewidth]{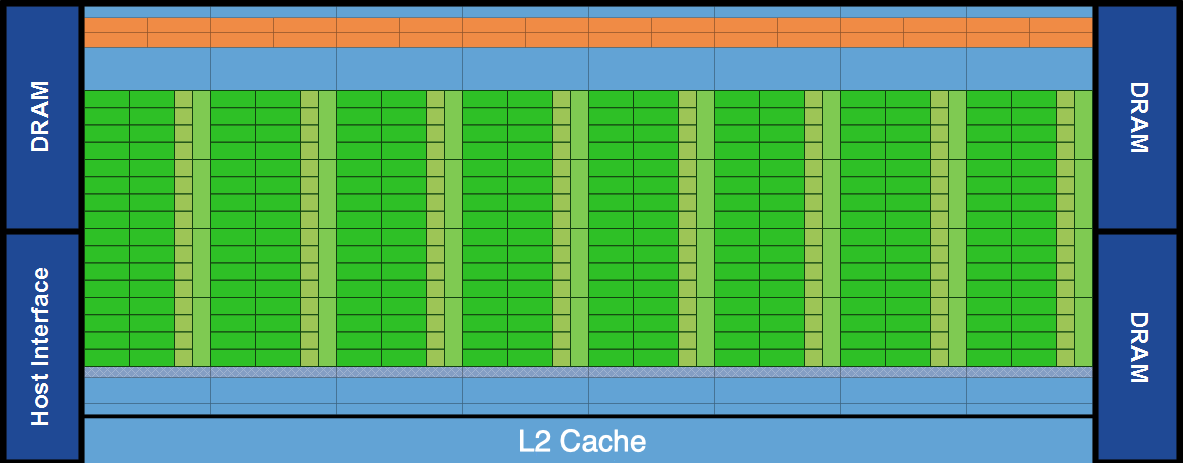}
        \caption{Eight (of 16) SM in the Fermi architecture (remaining 8 are symmetrically placed around the L2 cache)}
        \label{fig:top_half_ferm_arch}
    \end{subfigure}
    \\[3ex]
    \begin{subfigure}{\linewidth}
        \centering
        \includegraphics[width=\linewidth]{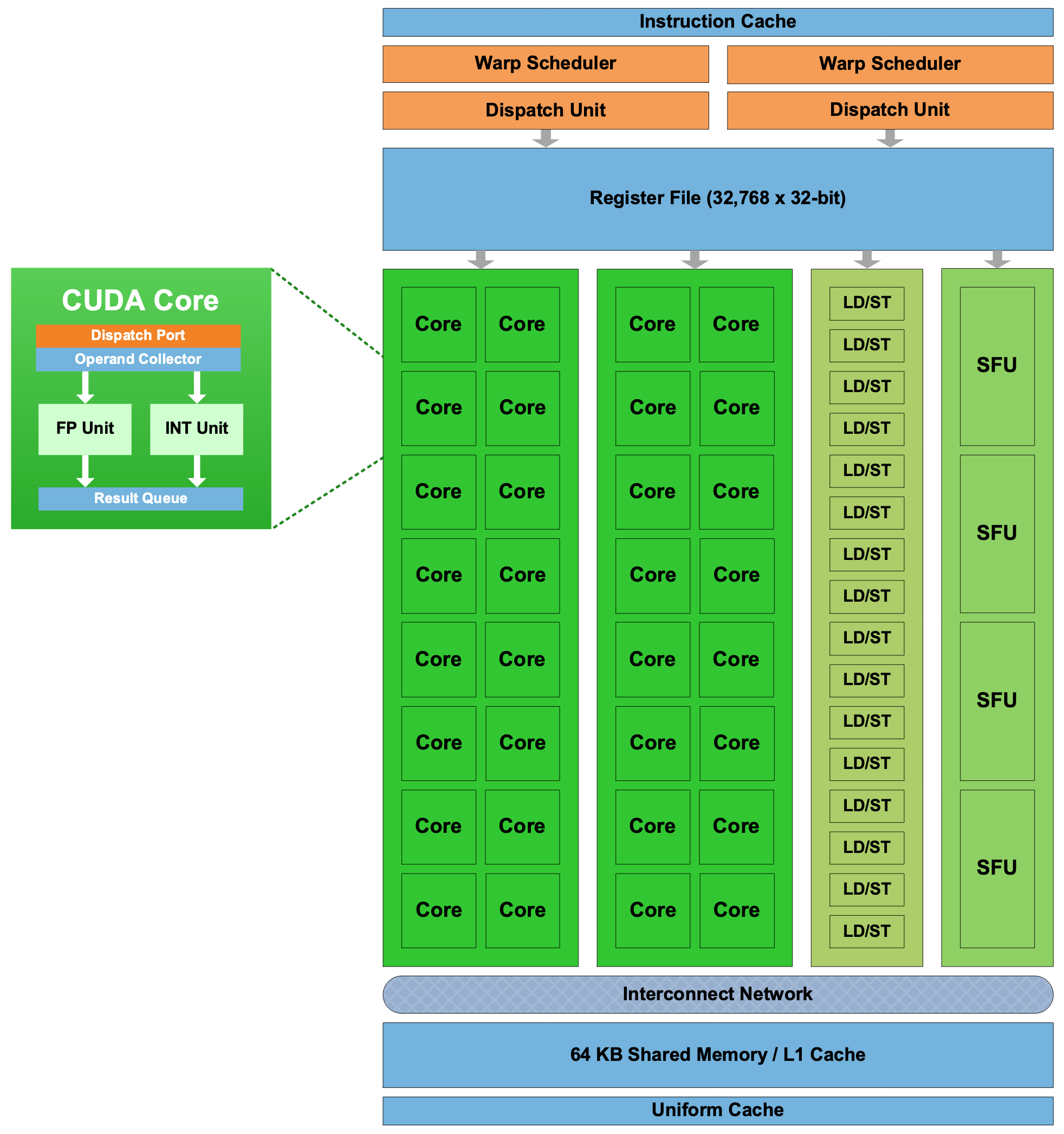}
        \caption{An individual Fermi SM}
        \label{fig:cuda_cores}
    \end{subfigure}
    \caption{NVIDIA Fermi Architecture~\cite{5751939}}
    \label{fig:fermi_arch}
\end{figure}

A GPU consists of many simple processors, called streaming multiprocessors (SMs), which are comprised by many compute \textit{cores} that run at relatively low clock speeds%
\footnote{For example, individual NVIDIA GTX-1080 Ti cores run at $\sim$1500MHz.}.
Each compute core in an SM can execute one floating-point or integer operation per clock cycle.
See ~\cref{fig:fermi_arch} for a diagram of NVIDIA's Fermi architecture, where each SM consists of 32 cores, 16 load/store (LD/ST) units, four special-function units (SFUs) which compute transcendental functions (such as $\sin$, $\cos$, $\exp$), a relatively large register file%
\footnote{For example, Intel's Haswell architecture supports 168 integer and 168 floating-point registers.}%
, and thread control logic (to be discussed in the proceeding).
Each SM has access to local memory, several cache levels, and global memory.
In the Fermi architecture (and subsequent architectures) local memory is configurable in software;
a fraction of it can be apportioned as either local memory or L1 cache (for workloads that query global memory in excess of local memory).
One final feature worth mentioning, though irrelevant for us here, is the L2 cache's atomic \code{read-modify-write} facilities;
this enables sharing data across groups of threads more efficiently than possible in conventional CPUs%
\footnote{On a CPU, atomic \code{test-and-set} instructions manage a semaphore, which itself manages access to memory (therefore incurring a cost of at least two clock cycles).}.

Such an architecture, particularly suited to maximizing throughput, necessitates a programming model distinct from that of a conventional, general purpose processor architecture.
A unit of computation deployed to a GPU is called a \textit{kernel}; kernels can be defined using NVIDIA's Compute Unified Device Architecture (CUDA) extensions to C, C++, and FORTRAN%
\footnote{In fact, CUDA compiles down to a virtual machine assembly code (by way of \code{nvcc}) for a virtual machine called the Parallel Thread Execution (PTX) virtual machine. So, in effect, it is compilers all the way down.}.
Compiled kernels are executed by many \textit{threads} in parallel, with each thread starting at the same instruction;
NVIDIA describes this addition to Flynn's taxonomy~\cite{5009071} as Single Instruction Multiple Thread (SIMT)%
\footnote{They key difference between SIMD and SIMT is that while in SIMD all vector elements in a vector instruction execute synchronously, threads in SIMT can diverge; branches are handled by predicated instructions~\cite{cuda_toolkit}.}.
The large register file enables very fast thread context switching ($\sim$25 microseconds on the Fermi architecture~\cite{Glaskowsky2009NVIDIAS}), performed by a centralized hardware thread scheduler.
Multiple threads are grouped into blocks (SMs are single tenant with respect to blocks) and blocks are grouped into \textit{grids} (grids execute a single kernel).
All threads in a block, by virtue of running on the same SM, coordinate (execute in arbitrary order, concurrently, or sequentially) and share memory.
Thread blocks are partitioned into \textit{warps} of 32 threads;
it is these warps that are dispatched by the warp scheduler (see~\cref{fig:cuda_cores}) and starting with the Fermi architecture two warps can be executed concurrently on the same SM in order to increase utilization%
\footnote{That is, one warp can occupy the compute cores while the other occupies the SFUs or Load/Store units.}.

\begin{figure}
    \centering
    \begin{subfigure}{\linewidth}
        \centering
        \cfile{code/hello_world_kernel.c}
        \caption{CUDA code to be compiled by \mintinline{c}{nvcc}.
        Note differences \mintinline{c}{__global__} and \mintinline{c}{matrix_sum<<,>>} from standard C.
        }
        \label{lst:cuda_hello_world}
    \end{subfigure}
    \\[3ex]
    \begin{subfigure}{\linewidth}
        \centering
        \includegraphics[width=\linewidth]{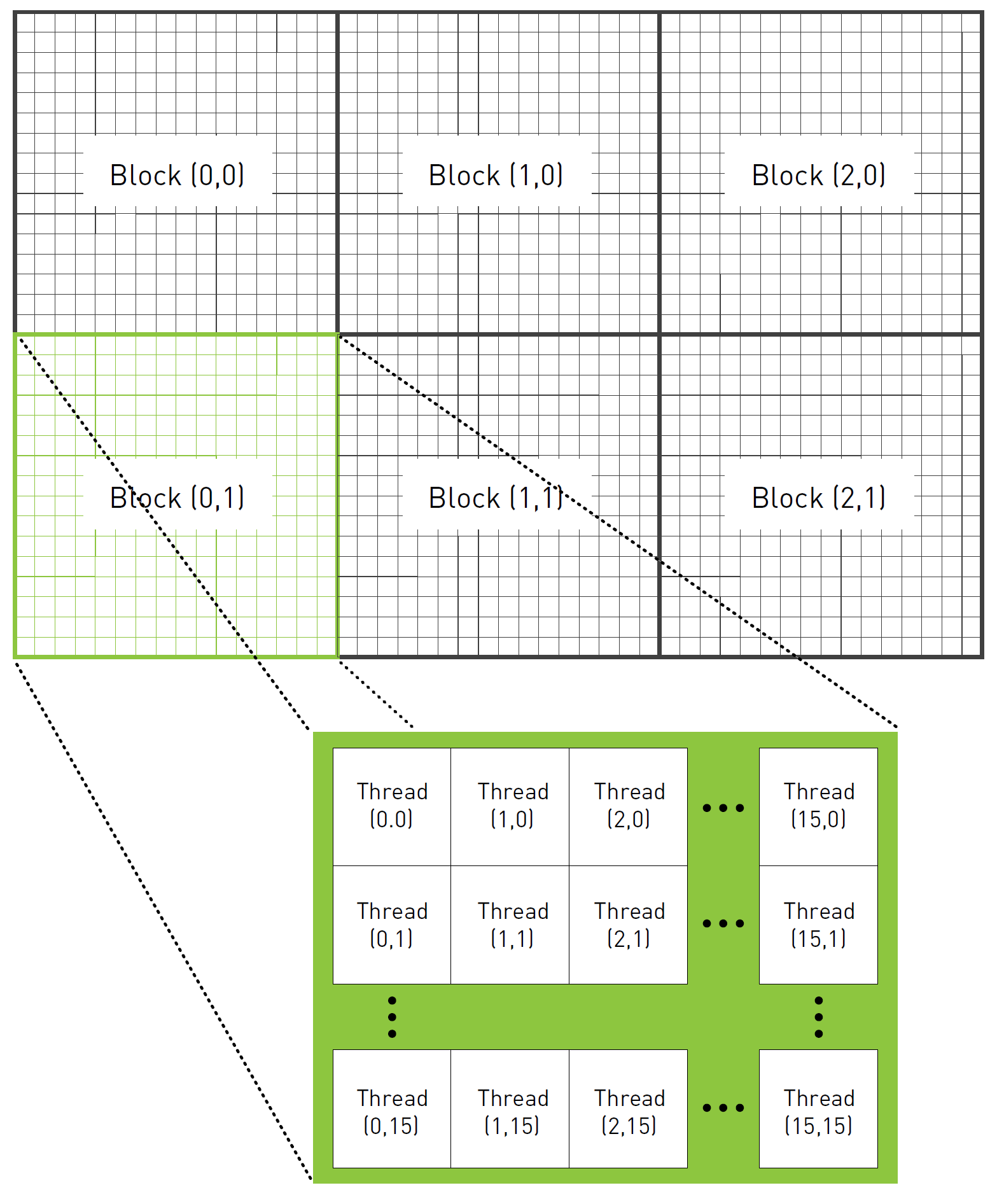}
        \caption{Mapping from thread and block to matrix element~\cite{10.5555/1891996}.}
        \label{fig:matrix_thread}
    \end{subfigure}
    \caption{Canonical CUDA "hello world" kernel (matrix addition).}
    \label{fig:cuda_hello_world}
\end{figure}

We present an example CUDA program in~\cref{fig:cuda_hello_world} to illustrate some of the artifacts of the CUDA threading model.
The premise of the program is performing an element-wise sum of two $32 \times 48$ entry matrices.
Note that all of the data weighs in at  $3 \times 32 \times 48 \times 4 = 18$ kilobytes (well within the bounds of shared memory on any one SM).
The actual work of summing is partitioned across a grid of six thread blocks, each containing $16 \times 16$ threads.
Such a partitioning means each thread can be logically responsible for exactly one sum and therefore the kernel is quite simple (see~\cref{lst:cuda_hello_world}).
Within the context of a kernel, each thread is uniquely identified by its multi-index in the thread hierarchy (\code{threadIdx} and \code{blockIdx}).
Hence, to carry out the sum, the kernel maps this multi-index to the physical address of the data%
\footnote{In CUDA C/C++ data is laid out in row-major order but this is not fixed (in CUDA FORTRAN the data is laid out in column-major order).}.
This (grid, block, thread)-to-data mapping is, in effect, the mechanism that implements the SIMT architecture.
Note that, since each block is allocated to exactly one SM, this sum will take $\left( 16 \times 16 \right) \div 16 = 16$ clock cycles on the Fermi architecture;
better throughput could be achieved by increasing the number of blocks (and therefore the number of SMs assigned work).

\subsection{Graph compilers and Tensors}\label{subsec:graph-compilers}

DL frameworks primarily function as graph compilers and tensor abstractions\footnote{A tensor in this context is a data structure similar to a multidimensional array that supports some useful operations (e.g. slicing, flattening, index permutation). Most DL frameworks also abstract memory layout on hardware behind this abstraction.}.
They typically also include some ``quality of life'' utilities useful for the training of DL models (e.g.\ optimizers and data loaders).
PyTorch's \code{Tensor} abstraction is responsible for a great deal of the complexity and implementation overhead of the framework.
Due to the framework's broad support for hardware and data types, dynamic dispatch\footnote{Kernels live in shared-object libraries (e.g. \code{libcaffe2.so}, \code{libcaffe2\_gpu.so}) and therefore call sites of virtual functions (indirection) are resolved at runtime.} is employed to resolve methods on \code{Tensor}s (see~\cref{fig:dispatch}).
This dynamic dispatch produces deep call stacks for every single operation on a \code{Tensor} (see~\cref{fig:stacks}); it remains to be seen whether the context switching\footnote{Every function call corresponds to a stack frame allocation and register allocations. In addition indirection to far away call sites leads to poor instruction cache efficiency~\cite{10.5555/3314872.3314876}} between function contexts incurs any appreciable execution time penalty.

\begin{figure*}[h]
    \begin{tikzpicture}[>=latex]
        \tikzset{block/.style= {draw,rectangle,align=center,minimum width=2cm,minimum height=1cm}}
        \node [](code)  {\code{torch.conv2d(x,y)}};
        \node [above right =1.25cm and 2cm of code, text=red](CUDA)   {\code{CUDA}};
        \node [below=0cm of CUDA, text=red](CPU) {\code{CPU}};
        \node [below=0cm of CUDA, text=red](CPU) {\code{CPU}};
        \node [below=0cm of CPU, text=red](FPGA) {\code{FPGA}};
        \node [below=0cm of FPGA, text=red](XLA) {\code{XLA}};
        \node [below=0cm of XLA, text=red](dots) {\vdots};
        \node [below=0cm of dots, text=red](vulkan) {\code{Vulkan}};
        \node [below=0cm of vulkan, text=red](mkldnn) {\code{MKLDNN}};
        \node [below=0cm of mkldnn, text=red](qcuda) {\code{QuantizedCUDA}};
        \node[block,draw=red,dashed, text=red, inner sep=5mm, fit= (CUDA) (qcuda),label={[anchor=south,text=red]north:dynamic dispatch}](dynamic) {};

        \node [right=2cm of CPU](SparseCUDAByteType) {\code{SparseCUDAByteType::conv2d}};
        \node [below=0cm of SparseCUDAByteType](twodots) {\vdots};
        \node [below=0cm of twodots](CUDADoubleType) {\code{CUDADoubleType::conv2d}};
        \node [below=0cm of CUDADoubleType](threedots) {\vdots};
        \node [below=0cm of threedots](CUDAHalfType) {\code{CUDAHalfType::conv2d}};
        \node[block,draw=black,dashed, inner sep=5mm, fit= (SparseCUDAByteType) (CUDAHalfType),label={[anchor=south]north:static\ dispatch\ on\ data\ type}](dtype) {};

        \node [right=2cm of CUDADoubleType](cudnnforward) {\code{cudnnConvolutionForward}};

        \path[draw]
        (code.east) edge[->] (dynamic.west);
        \path[draw]
        (CUDA.east) edge[out=0, in=180,->] (SparseCUDAByteType.west);
        \path[draw]
        (CUDA.east) edge[out=0, in=180,->] (CUDADoubleType.west);
        \path[draw]
        (CUDA.east) edge[out=0, in=180,->] (CUDAHalfType.west);
        \path[draw]
        (dtype.east) edge[out=0, in=180,->] (cudnnforward.west);
    \end{tikzpicture}
    \caption{How the \code{torch.conv2d} operation on tensors \code{x}, \code{y} is implemented in PyTorch.}\label{fig:dispatch}
\end{figure*}
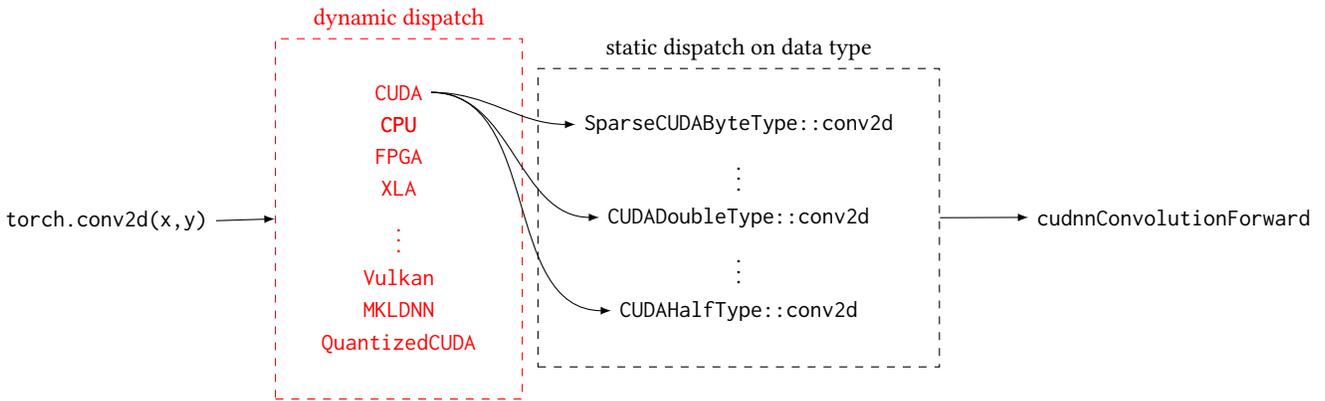

\begin{figure*}[h]
\centering
\begin{adjustbox}{width=0.8\textwidth}
  \centering
  \begin{tikzpicture}[%
      >=latex,
      level distance=10mm,
      inner sep=2mm,
      edge from parent path={[thick,->] ([thick]\tikzparentnode.south) -> (\tikzchildnode.north)}]
    \tikzstyle{calls}=[draw=black,thick,anchor=west]
    \tikzstyle{commoncode}=[pattern=north west lines, pattern color=green!50, draw=green!50!black]
    \tikzstyle{uncommoncode}=[pattern=north west lines, pattern color=red!50, draw=red!50!black]
    \node[] (node) at (1.2,0) {\Large PyTorch};
    \node[] (node) at (4.5,-2) {\Large LibTorch};
    \node[] (node) at (6.5,-7.2) {\Large cuDNN};
    \node[] (node) at (9.5,-5) {\Large TorchScript};
    \node[] (node) at (0,0) {}
    child[] { node[calls,uncommoncode] { \code{torch.conv2d} } edge from parent[draw=none]
        child { node[uncommoncode] {$\bigcdot$}
            child { node[uncommoncode] {$\bigcdot$} edge from parent[draw=none]
                child { node[uncommoncode] {$\bigcdot$} edge from parent[draw=none]
                    child[] { node[calls,commoncode] {\code{at::conv2d}}
                        child[grow=up, edge from parent path={[<-] ([thick]\tikzparentnode.north) -> (\tikzchildnode.south)}] { node[calls, uncommoncode] {\code{functional::detail::conv2d}}
                            child[grow via three points={one child at (-1.72,1) and two children at (0.1,-1) and (0.1,-2)}, edge from parent path={[<-] ([]\tikzparentnode.north) -> (\tikzchildnode.south)}] { node[calls, uncommoncode] {\code{Conv2dImpl::forward}}
                              }
                          }
                        child { node[commoncode] {$\bigcdot$}
                            child { node[commoncode] {$\bigcdot$} edge from parent[draw=none]
                                child { node[commoncode] {$\bigcdot$} edge from parent[draw=none]
                                    child[grow via three points={one child at (-0.5,-1) and
                                            two children at (0,1) and (5,0)}] { node[calls,commoncode] {\code{cudnnConvolutionForward}}
                                        child[edge from parent path={[<-] ([]\tikzparentnode.north) -> (\tikzchildnode.south)}] { node[calls, uncommoncode] {\code{Conv2d<float>::forward}}
                                          }
                                        child[edge from parent path={[<-] ([]\tikzparentnode.east) -> (\tikzchildnode.west)}] { node[uncommoncode] {$\bigcdot$}
                                            child[grow=up, edge from parent path={[<-] ([]\tikzparentnode.east) -> (\tikzchildnode.west)}] { node[uncommoncode] {$\bigcdot$} edge from parent[draw=none]
                                                child[grow=up, edge from parent path={[<-] ([]\tikzparentnode.east) -> (\tikzchildnode.west)}] { node[uncommoncode] {$\bigcdot$} edge from parent[draw=none]
                                                    child[grow via three points={one child at (-3.07,1) and
                                                            two children at (0,0) and (0,0)}, edge from parent path={[<-] ([]\tikzparentnode.north) -> (\tikzchildnode.south)}] { node[calls, uncommoncode] {\code{autograd::VariableType::\_convolution}}
                                                      }
                                                  }
                                              }
                                          }
                                      }
                                  }
                              }}}}}}};
    \draw[thick,|<->|]
    (0.22,-2) -- (0.22,-4) node [midway, fill=white] {30 calls};
    \draw[thick,|<->|]
    (1.9,-6) -- (1.9,-8) node [midway, fill=white] {61 calls};
    \draw[thick,|<->|]
    (10.5,-7) -- (10.5,-9) node [midway, fill=white] {14 calls};
  \end{tikzpicture}
\end{adjustbox}
      \caption["Short" caption without tikz code]{Call graphs representing the number of calls between \code{Conv2d.forward} at the level of abstraction and the ultimate execution of the convolution \code{cudnnConvolutionForward} on the GPU. \tikz{\path[fill=red!10,draw=red!50!black] (0,0) rectangle (.25cm,.25cm);}\, represents calls where the implementations diverge and \tikz{\path[pattern=north west lines, pattern color=green!50, draw=green!50!black] (0,0) rectangle (.25cm,.25cm);} represents calls where two or more implementations coincide. Note that program setup calls are omitted. These were produced by building each implementation with debug symbols and using \code{gdb} to set a breakpoint at \code{cudnnConvolutionForward}. Complete stacktraces are available on GitHub at \link{\href{https://github.com/makslevental/pytorch_abstraction_comparison/tree/main/tex/figures/stack_traces}{main/tex/stack\_traces}}.}
  \label{fig:stacks}
\end{figure*}

DL graph compilers are distinct from other dataflow compilers (such as VHDL and Verilog\footnote{Verilog and Very High Speed Integrated Circuit Hardware Description Language (VHSIC-HDL or VHDL) are specification languages for specifying circuits on field programmable gate arrays.}); in addition to keeping account of how the data streams through the compute graph, they also keep account of how the gradients of the data stream through the graph (i.e.\ the \textit{gradient-flow}).
This is called \textit{automatic differentiation} (often shortened to \textit{autodiff}).
In principle autodiff is implemented by using the rules of Newton's calculus to calculate the derivatives of primitive functions and the chain rule to calculate derivatives of compositions of primitive functions.
There are two types of autodiff: \textit{forward mode} (or \textit{forward accumulation}) and \textit{reverse mode} (or \textit{reverse accumulation})%
\footnote{Briefly, for a composition of functions $y=f(g(h(x)))$, forward mode evaluates the derivative $y'(x)$, as given by the chain rule, inside-out while reverse mode evaluates the derivative outside-in. For those familiar with functional programming, these operations correspond to \code{foldl} and \code{foldr} on the sequence of functions with $\partial_x$ as the operator.}.
Reverse mode autodiff enables the framework to effectively calculate the gradients of parameters of a neural network with respect to some relevant loss or objective function.
Note that such gradients can be \textit{back-propagated} through the neural network in order to adjust the parameters of the neural network such that it minimizes the loss\footnote{In which case, it is, in fact, the negatives of the gradients that are back-propagated.} or maximizes the objective.

Dataflow graphs (and their corresponding gradient-flow graphs) can be specified either statically, with fan-in and fan-out for all functions predetermined, or dynamically, where compositions of functions are determined ``on-the-run''.
There are advantages and disadvantages to both specification strategies.
Static specifications tightly constrain\footnote{For example, branches and loops are cumbersome to specify statically.} the intricacy of the dataflow graph but, obversely, can be leveraged to improve performance and scalability~\cite{le2019tflms,Pradelle2017PolyhedralOO}.
TensorFlow (prior to v2.0) is an example of a DL framework that compiles statically specified graphs.
Conversely, dynamic specifications can be very expressive and user friendly, including such conveniences as runtime debugging, but are much more difficult to optimize.
PyTorch is an example of a DL framework that supports dynamic specification.
Both PyTorch and TensorFlow also support just-in-time (JIT) compilation strategies (TorchScript and XLA respectively);
such JIT compilers strike a balance between fluency and scalability.
In this work we investigate TorchScript (see~\cref{sec:methodology}).

It warrants mention that, in addition to vertically integrated DL frameworks (i.e.\ specification language and hardware compiler), recently there has been work on intermediate bytecode representations for dataflow graphs that arbitrary compiler ``frontends'' can target.
The Multi-Level Intermediate Representation (MLIR)~\cite{lattner2020mlir} project has goals that include supporting dataflow graphs, optimization passes on those graphs and hardware specific optimizations%
\footnote{Interestingly enough, the project is headed by Chris Lattner who, in developing LLVM, pioneered the same ideas in general purpose programming languages.}.
Stripe~\cite{zerrell2019stripe} is a polyhedral compiler%
\footnote{A polyhedral compiler models complex programs (usually deeply nested loops) as polyhedra and then performs transformations on those polyhedra in order to produce equivalent but optimized programs~\cite{Griebl98codegeneration}.}
that aims to support general machine learning kernels, which are distinguished by their high parallelism with limited mutual dependence between iterations.
Tensor Comprehensions~\cite{vasilache2018tensor} is an intermediate specification language (rather than intermediate bytecode representation) and corresponding polyhedral compiler;
the syntax bears close resemblance to Einstein summation notation and the compiler supports operator fusion and specialization for particular data shapes.
Finally, Tensor Virtual Machine (TVM)~\cite{10.5555/3291168.3291211} is an optimizing graph compiler that automates optimization using a learning-based cost modeling method that enables it to efficiently explore the space of low-level code optimizations.

\section{Methods}\label{sec:methodology}
As discussed in the preceding, the translation from high-level neural network specification to native hardware involves a diverse set of choices at design time and translation time.
Any such choice made implicitly by the DL framework abstracts away intermediary details at the level of abstraction at which the choice is made.
For example, in PyTorch, by default, convolutions include not only learnable filters but also a learnable bias.
Naturally this increases the number of parameters for which gradients need to be kept account of and updated for.
At the next level of abstraction (translation from Python specification to C++ objects) another implicit choice is made in choosing tensor dimension ordering%
\footnote{PyTorch uses some heuristics to order tensor dimensions as either NCHW or NHWC.}.
Finally, at the level of abstraction just prior to compilation into CUDA kernels a convolution strategy is chosen%
\footnote{Winograd convolution, general matrix multiply (GEMM), or FFT convolution.} according to heuristics.
Each of these choices potentially incurs a runtime execution and memory cost, depending on whether the heuristics according to which the choice was made apply to the user's DL model.

\begin{figure}
    \includegraphics[width=.6\linewidth]{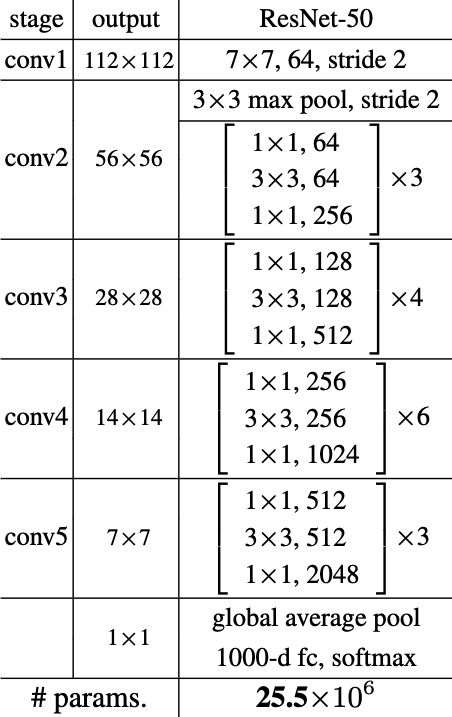}
    \caption{ResNet-50 network architecture~\cite{he2015deep}. Note that each convolution is followed by a batch normalization unit and each ``stage'' is followed by a ReLU (residual connections omitted).}
    \label{fig:resnet}
\end{figure}

With this as subtext, we describe the intent and design of our experiments.
We implement the popular object detection deep neural network ResNet-50~\cite{he2015deep} at four levels of abstraction (PyTorch, TorchScript\footnote{TorchScript models are serializations of PyTorch models but can run in inference mode in C++, i.e. sans Python runtime.}, LibTorch, and cuDNN) in order to investigate the differences amongst them.
We measure accuracy, execution time, GPU utilization, and memory efficiency of each implementation on four image datasets (MNIST, CIFAR10, STL10, PASCAL).
The source for the implementations is available on GitHub\footnote{\link{\url{https://github.com/makslevental/pytorch_abstraction_comparison}}}.
The datasets were chosen because they span the spectrum of image complexity (from small single-channel images to large multi-channel images).
The reasons for choosing ResNet-50 (see~\cref{fig:resnet}) are two fold.
Firstly, it serves as a benchmark architecture in the research community.
Secondly, it includes functional units included in many other network architectures (residual units, convolutions of various sizes, batch normalizations, ReLU activations, and pooling layers) and is therefore representative of typical neural network compute workloads.
The reason for staying within the same ecosystem is that, in theory, we fix as many of the dimensions of functionality orthogonal to our concerns as possible.

We employ two test platforms (see~\cref{tab:platforms});
training is performed for 100 epochs with batch size 128 (except for on PASCAL where we use batch size 16) on the ``Training platform''.
Distinct models were trained in parallel in order to expedite the overall training process but neither data parallelism nor intra-model parallelism were employed.
In addition we perform scaling experiments (in resolution and batch size) on the PASCAL dataset;
for $\code{batch\_size}, \code{resolution} = 8, 16, \dots, 1024$.
For this purpose we employ the ``Resolutions platform'', which has GPU RAM enough to accommodate large batch sizes and large image resolutions.
For both sets of experiments, each run is repeated 10 times and results are averaged to reduce variance in the measurements.
Precise execution time measurements are collected using the CUDA \code{cudaEventCreate}, \code{cudaEventRecord}, \code{cudaEventElapsedTime} APIs.
Precise memory and GPU utilization measurements are collected using the NVIDIA Management Library C API\@.
Both sets of APIs were appropriately wrapped for use in Python.

\begin{table}
  \caption{Test platforms}
  \centering
  \begin{tabular}[t]{p{0.15\linewidth}p{0.75\linewidth}}
    \hline
    CPU      & AMD Ryzen 2970WX 24-Core @ 4.2 GHz                              \\
    GPU      & 4 $\times$ GeForce GTX 1080Ti                                              \\
    HD       & Crucial MX500 2TB 3D NAND SATA                                  \\
    RAM      & 64GB                                                            \\
    Software & PyTorch-1.7.0, CUDA-11.1, NVIDIA-455.23.05, g++10.1.0           \\
    \hline
  \end{tabular}
  \subcaption*{Training platform}
  \bigskip
  \begin{tabular}[t]{p{0.15\linewidth}p{0.75\linewidth}}
    \hline
    CPU      & Intel Xeon Gold 6230 CPU @ 2.10GHz                              \\
    GPU      & Tesla V100-PCIE-32GB                                            \\
    HD       & HPE 800GB SAS 12G Mixed Use SFF                                 \\
    RAM      & 384GB                                                           \\
    Software & PyTorch-1.7.0, CUDA-11.1, NVIDIA-450.80.02                      \\
    \hline
  \end{tabular}
  \subcaption*{Resolutions platform}\label{tab:platforms}
\end{table}

\section{Results}\label{sec:results}
\begin{figure*}
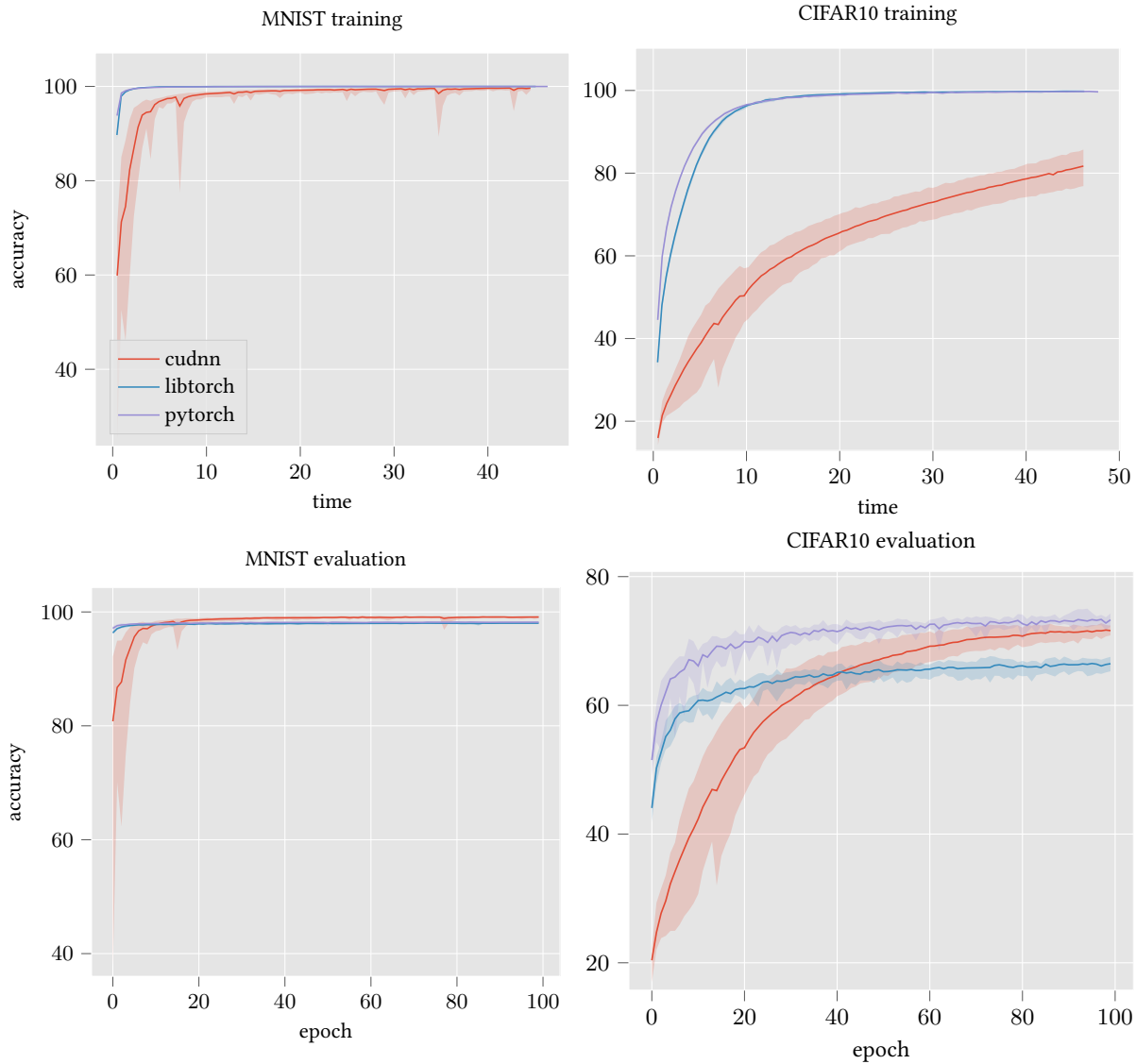

    \centering
    \begin{subfigure}{0.45\linewidth}
        \centering
        \begin{adjustbox}{width=\textwidth}
            \input{figures/plots/train_accuracy_per_epoch_mnist.tex}
        \end{adjustbox}
    \end{subfigure}%
    \begin{subfigure}{0.45\textwidth}
        \centering
        \begin{adjustbox}{width=\textwidth}
            \input{figures/plots/train_accuracy_per_epoch_cifar10.tex}
        \end{adjustbox}
    \end{subfigure}
    \quad
    \begin{subfigure}{0.45\textwidth}
        \centering
        \begin{adjustbox}{width=\textwidth}
            \input{figures/plots/eval_accuracy_per_epoch_mnist.tex}
        \end{adjustbox}
    \end{subfigure}
    \begin{subfigure}{0.45\textwidth}
        \centering
        \begin{adjustbox}{width=\textwidth}
            \input{figures/plots/eval_accuracy_per_epoch_cifar10.tex}
        \end{adjustbox}
    \end{subfigure}
    \caption{Comparison of accuracy for PyTorch, LibTorch, and cuDNN implementations during training and evaluation. Solid line corresponds to mean while shaded regions correspond to min and max. Time is measured in units of \textit{epoch} $\times$ \textit{average epoch time}. }
    \label{fig:accuracy_results}
\end{figure*}

The PyTorch implementation compares favorably with both LibTorch and the cuDNN implementations (see~\cref{fig:accuracy_results}) in terms of accuracy.
On MNIST and CIFAR10 all three implementations perform reasonably well;
LibTorch and PyTorch attain maximum accuracy at around the same time while cuDNN lags behind.
On the more complex STL10 and PASCAL datasets (see~\cref{fig:other_accuracy_results} in the appendix) the cuDNN implementation dramatically underperformed PyTorch and LibTorch.
The cause of the difference between the cuDNN implementation and the others is unclear.

In attempting to resolve the poor performance of the cuDNN implementation it was discovered that PyTorch (and LibTorch as well) initializes weights in convolutional, linear, and batch normalization layers.
This is not documented and not configurable.
For convolutional and linear layers Kaiming uniform initialization~\cite{he2015delving} is used and for batch normalization layers $(\gamma=1,\beta=0)$ initialization is used.
This presents a serious problem for us because it is known that \textbf{ResNets with Kaiming initializations lead to exploding gradients}~\cite{zhang2019fixup}.
Nonetheless, we implemented Kaiming initialization for the cuDNN implementation but it did not resolve the under-performance issues.
Indeed, the network vacillated between vanishing gradients and exploding gradients depending on various settings of the hyper-parameters in the Kaiming initialization.
Note that TorchScript training and evaluation accuracy is not measured/reported because TorchScript implementations cannot (as of yet) be trained, only evaluated.

The undocumented initialization leads us to believe that most likely there are several other heuristic optimizations implemented by Pytorch (and LibTorch).
While such optimizations generally do improve performance (to wit: here on STL10 and PASCAL) this prompts the question of whether or not this is a ``moral'' cost of abstraction (since the optimizations might hurt performance for other models~\cite{zhang2019fixup}).

\begin{figure*}
    \centering
    \begin{subfigure}{0.45\linewidth}
        \centering
        \begin{adjustbox}{width=\textwidth}
\begin{tikzpicture}

\begin{axis}[
axis background/.style={fill=white!89.8039215686275!black},
axis line style={white},
legend cell align={left},
legend style={fill opacity=0.8, draw opacity=1, text opacity=1, at={(0.03,0.97)}, anchor=north west, draw=white!80!black, fill=white!89.8039215686275!black},
log basis y={10},
tick align=outside,
tick pos=left,
title={Average sample time for batch size $=32$},
x grid style={white},
xlabel={resolution},
xmajorgrids,
xmin=2.65, xmax=10.35,
xtick style={color=white!33.3333333333333!black},
y grid style={white},
ylabel={time [ms]},
ymajorgrids,
ymin=0.152640394088257, ymax=27.8442975883733,
ymode=log,
ytick style={color=white!33.3333333333333!black},
xticklabels={$2^3$, $2^4$, $2^5$, $2^6$, $2^7$, $2^8$, $2^9$, $2^{10}$},
xtick={3,...,10},
]
\path [fill=cudnn, fill opacity=0.2, very thin]
(axis cs:3,0.201026)
--(axis cs:3,0.19947)
--(axis cs:4,0.199807)
--(axis cs:5,0.227339)
--(axis cs:6,0.330417)
--(axis cs:7,0.861158)
--(axis cs:8,2.71924)
--(axis cs:8,2.77216)
--(axis cs:8,2.77216)
--(axis cs:7,0.88903)
--(axis cs:6,0.3433)
--(axis cs:5,0.230137)
--(axis cs:4,0.204628)
--(axis cs:3,0.201026)
--cycle;

\path [fill=libtorch, fill opacity=0.2, very thin]
(axis cs:3,0.239017)
--(axis cs:3,0.224985)
--(axis cs:4,0.223488)
--(axis cs:5,0.227935)
--(axis cs:6,0.260423)
--(axis cs:7,0.475671)
--(axis cs:8,1.50341)
--(axis cs:9,5.67418)
--(axis cs:9,5.86691)
--(axis cs:9,5.86691)
--(axis cs:8,1.56544)
--(axis cs:7,0.483492)
--(axis cs:6,0.261208)
--(axis cs:5,0.242581)
--(axis cs:4,0.238601)
--(axis cs:3,0.239017)
--cycle;

\path [fill=pytorch, fill opacity=0.2, very thin]
(axis cs:3,1.264947)
--(axis cs:3,1.21564)
--(axis cs:4,1.213545)
--(axis cs:5,1.215587)
--(axis cs:6,1.220171)
--(axis cs:7,1.296469)
--(axis cs:8,2.795173)
--(axis cs:9,9.585992)
--(axis cs:9,9.846071)
--(axis cs:9,9.846071)
--(axis cs:8,2.891976)
--(axis cs:7,1.341868)
--(axis cs:6,1.291262)
--(axis cs:5,1.29437)
--(axis cs:4,1.259168)
--(axis cs:3,1.264947)
--cycle;

\path [fill=white!46.6666666666667!black, fill opacity=0.2, very thin]
(axis cs:3,0.353049)
--(axis cs:3,0.199058)
--(axis cs:4,0.193395)
--(axis cs:5,0.195655)
--(axis cs:6,0.227985)
--(axis cs:7,0.435408)
--(axis cs:8,1.42414)
--(axis cs:9,5.45931)
--(axis cs:10,20.8873)
--(axis cs:10,21.9766)
--(axis cs:10,21.9766)
--(axis cs:9,5.7139)
--(axis cs:8,1.5859)
--(axis cs:7,0.616575)
--(axis cs:6,0.401246)
--(axis cs:5,0.348112)
--(axis cs:4,0.374466)
--(axis cs:3,0.353049)
--cycle;

\path [fill=torchscript, fill opacity=0.2, very thin]
(axis cs:3,0.353049)
--(axis cs:3,0.199058)
--(axis cs:4,0.193395)
--(axis cs:5,0.195655)
--(axis cs:6,0.227985)
--(axis cs:7,0.435408)
--(axis cs:8,1.42414)
--(axis cs:9,5.45931)
--(axis cs:10,20.8873)
--(axis cs:10,21.9766)
--(axis cs:10,21.9766)
--(axis cs:9,5.7139)
--(axis cs:8,1.5859)
--(axis cs:7,0.616575)
--(axis cs:6,0.401246)
--(axis cs:5,0.348112)
--(axis cs:4,0.374466)
--(axis cs:3,0.353049)
--cycle;

\addplot [semithick, cudnn]
table {%
3 0.200238913297653
4 0.202598989009857
5 0.229123681783676
6 0.338942557573318
7 0.870264947414398
8 2.73578190803528
};
\addlegendentry{cudnn}
\addplot [semithick, libtorch]
table {%
3 0.232446998357773
4 0.230680495500565
5 0.237425118684769
6 0.260794192552567
7 0.47879546880722
8 1.53802895545959
9 5.7626690864563
};
\addlegendentry{libtorch}
\addplot [semithick, pytorch]
table {%
3 1.23768603801727
4 1.23296797275543
5 1.24600088596344
6 1.25468814373016
7 1.31825852394104
8 2.83480596542358
9 9.69671154022217
};
\addlegendentry{pytorch}
\addplot [semithick, torchscript]
table {%
3 0.220640391111374
4 0.216945394873619
5 0.211528897285461
6 0.245379000902176
7 0.460912704467773
8 1.45570909976959
9 5.53999090194702
10 21.3651218414307
};
\addlegendentry{torchscript}
\end{axis}

\end{tikzpicture}
        \end{adjustbox}
        \label{fig:avg_sample_time_32}
    \end{subfigure}%
    \begin{subfigure}{0.45\textwidth}
        \centering
        \begin{adjustbox}{width=\textwidth}
\begin{tikzpicture}

\begin{axis}[
axis background/.style={fill=white!89.8039215686275!black},
axis line style={white},
log basis y={10},
tick align=outside,
tick pos=left,
title={Average sample time for resolution $=64$},
x grid style={white},
xlabel={batch size},
xmajorgrids,
xmin=2.7, xmax=9.3,
xtick style={color=white!33.3333333333333!black},
y grid style={white},
ymajorgrids,
ymin=0.0318818812180445, ymax=4.24381955156468,
ymode=log,
ytick style={color=white!33.3333333333333!black},
xticklabels={$2^3$, $2^4$, $2^5$, $2^6$, $2^7$, $2^8$, $2^9$},
xtick={3,...,9},
]
\path [fill=cudnn, fill opacity=0.2, very thin]
(axis cs:3,0.532711)
--(axis cs:3,0.525274)
--(axis cs:4,0.325467)
--(axis cs:5,0.227339)
--(axis cs:6,0.137164)
--(axis cs:7,0.127286)
--(axis cs:8,0.107328)
--(axis cs:9,0.0943382)
--(axis cs:9,0.100206)
--(axis cs:9,0.100206)
--(axis cs:8,0.1119)
--(axis cs:7,0.129387)
--(axis cs:6,0.145102)
--(axis cs:5,0.230137)
--(axis cs:4,0.333086)
--(axis cs:3,0.532711)
--cycle;

\path [fill=libtorch, fill opacity=0.2, very thin]
(axis cs:3,0.7206)
--(axis cs:3,0.643425)
--(axis cs:4,0.371694)
--(axis cs:5,0.227935)
--(axis cs:6,0.138768)
--(axis cs:7,0.110285)
--(axis cs:8,0.0866569)
--(axis cs:9,0.0769451)
--(axis cs:9,0.077837)
--(axis cs:9,0.077837)
--(axis cs:8,0.0870269)
--(axis cs:7,0.110558)
--(axis cs:6,0.149246)
--(axis cs:5,0.242581)
--(axis cs:4,0.39549)
--(axis cs:3,0.7206)
--cycle;

\path [fill=pytorch, fill opacity=0.2, very thin]
(axis cs:3,3.397831)
--(axis cs:3,3.123268)
--(axis cs:4,2.145072)
--(axis cs:5,1.215587)
--(axis cs:6,0.90215)
--(axis cs:7,0.489502)
--(axis cs:8,0.516421)
--(axis cs:9,0.290157)
--(axis cs:9,0.314908)
--(axis cs:9,0.314908)
--(axis cs:8,0.564356)
--(axis cs:7,0.522439)
--(axis cs:6,1.003318)
--(axis cs:5,1.29437)
--(axis cs:4,2.273831)
--(axis cs:3,3.397831)
--cycle;

\path [fill=white!46.6666666666667!black, fill opacity=0.2, very thin]
(axis cs:3,0.797093)
--(axis cs:3,0.633669)
--(axis cs:4,0.342584)
--(axis cs:5,0.195655)
--(axis cs:6,0.101828)
--(axis cs:7,0.071068)
--(axis cs:8,0.0474417)
--(axis cs:9,0.0398198)
--(axis cs:9,0.13526)
--(axis cs:9,0.13526)
--(axis cs:8,0.145412)
--(axis cs:7,0.180774)
--(axis cs:6,0.20295)
--(axis cs:5,0.348112)
--(axis cs:4,0.497252)
--(axis cs:3,0.797093)
--cycle;

\path [fill=torchscript, fill opacity=0.2, very thin]
(axis cs:3,0.797093)
--(axis cs:3,0.633669)
--(axis cs:4,0.342584)
--(axis cs:5,0.195655)
--(axis cs:6,0.101828)
--(axis cs:7,0.071068)
--(axis cs:8,0.0474417)
--(axis cs:9,0.0398198)
--(axis cs:9,0.13526)
--(axis cs:9,0.13526)
--(axis cs:8,0.145412)
--(axis cs:7,0.180774)
--(axis cs:6,0.20295)
--(axis cs:5,0.348112)
--(axis cs:4,0.497252)
--(axis cs:3,0.797093)
--cycle;

\addplot [semithick, cudnn]
table {%
3 0.527113080024719
4 0.328776597976685
5 0.229123681783676
6 0.139674320816994
7 0.128557801246643
8 0.10912749171257
9 0.098616324365139
};
\addplot [semithick, libtorch]
table {%
3 0.66579794883728
4 0.376946628093719
5 0.237425118684769
6 0.141885712742805
7 0.110373191535473
8 0.0867782235145569
9 0.077290840446949
};
\addplot [semithick, pytorch]
table {%
3 3.20224666595459
4 2.2035346031189
5 1.24600088596344
6 0.938888311386108
7 0.50709193944931
8 0.551432967185974
9 0.300852745771408
};
\addplot [semithick, torchscript]
table {%
3 0.662334620952606
4 0.360176265239716
5 0.211528897285461
6 0.112016521394253
7 0.0824229493737221
8 0.0573519766330719
9 0.0495683811604977
};
\end{axis}

\end{tikzpicture}
        \end{adjustbox}
    \end{subfigure}
    \\
    \begin{subfigure}{0.45\textwidth}
        \centering
        \begin{adjustbox}{width=\textwidth}
\begin{tikzpicture}

\begin{axis}[
axis background/.style={fill=white!89.8039215686275!black},
axis line style={white},
log basis y={10},
tick align=outside,
tick pos=left,
title={Average used memory for batch size $=32$},
x grid style={white},
xlabel={resolution},
xmajorgrids,
xmin=2.65, xmax=10.35,
xtick style={color=white!33.3333333333333!black},
y grid style={white},
ylabel={memory [MB]},
ymajorgrids,
ymin=981.538600682435, ymax=24927.29576095,
ymode=log,
ytick style={color=white!33.3333333333333!black},
xticklabels={$2^3$, $2^4$, $2^5$, $2^6$, $2^7$, $2^8$, $2^9$, $2^{10}$},
xtick={3,...,10},
]
\path [fill=cudnn, fill opacity=0.2, very thin]
(axis cs:3,1137)
--(axis cs:3,1137)
--(axis cs:4,1161)
--(axis cs:5,1291)
--(axis cs:6,1853)
--(axis cs:7,3819)
--(axis cs:8,11683)
--(axis cs:8,11683)
--(axis cs:8,11683)
--(axis cs:7,3819)
--(axis cs:6,1853)
--(axis cs:5,1291)
--(axis cs:4,1161)
--(axis cs:3,1137)
--cycle;

\path [fill=libtorch, fill opacity=0.2, very thin]
(axis cs:3,1559)
--(axis cs:3,1559)
--(axis cs:4,1567)
--(axis cs:5,1613)
--(axis cs:6,1785)
--(axis cs:7,2529)
--(axis cs:8,5591)
--(axis cs:9,18415)
--(axis cs:9,18417)
--(axis cs:9,18417)
--(axis cs:8,5591)
--(axis cs:7,2529)
--(axis cs:6,1785)
--(axis cs:5,1613)
--(axis cs:4,1567)
--(axis cs:3,1559)
--cycle;

\path [fill=pytorch, fill opacity=0.2, very thin]
(axis cs:3,1529)
--(axis cs:3,1529)
--(axis cs:4,1539)
--(axis cs:5,1583)
--(axis cs:6,1799)
--(axis cs:7,2667)
--(axis cs:8,6473)
--(axis cs:9,21519)
--(axis cs:9,21519)
--(axis cs:9,21519)
--(axis cs:8,6473)
--(axis cs:7,2667)
--(axis cs:6,1799)
--(axis cs:5,1583)
--(axis cs:4,1539)
--(axis cs:3,1529)
--cycle;

\path [fill=white!46.6666666666667!black, fill opacity=0.2, very thin]
(axis cs:3,1357)
--(axis cs:3,1357)
--(axis cs:4,1357)
--(axis cs:5,1359)
--(axis cs:6,1379)
--(axis cs:7,1459.02)
--(axis cs:8,1849.85)
--(axis cs:9,3454.26)
--(axis cs:10,9872.04)
--(axis cs:10,9891)
--(axis cs:10,9891)
--(axis cs:9,3459)
--(axis cs:8,1855)
--(axis cs:7,1461)
--(axis cs:6,1379)
--(axis cs:5,1359)
--(axis cs:4,1357)
--(axis cs:3,1357)
--cycle;

\path [fill=torchscript, fill opacity=0.2, very thin]
(axis cs:3,1357)
--(axis cs:3,1357)
--(axis cs:4,1357)
--(axis cs:5,1359)
--(axis cs:6,1379)
--(axis cs:7,1459.02)
--(axis cs:8,1849.85)
--(axis cs:9,3454.26)
--(axis cs:10,9872.04)
--(axis cs:10,9891)
--(axis cs:10,9891)
--(axis cs:9,3459)
--(axis cs:8,1855)
--(axis cs:7,1461)
--(axis cs:6,1379)
--(axis cs:5,1359)
--(axis cs:4,1357)
--(axis cs:3,1357)
--cycle;

\addplot [semithick, cudnn]
table {%
3 1136.99975585938
4 1161
5 1290.99975585938
6 1853
7 3819.00048828125
8 11683.0029296875
};
\addplot [semithick, libtorch]
table {%
3 1558.99975585938
4 1567.00024414062
5 1613.00024414062
6 1784.99975585938
7 2529.00024414062
8 5590.99951171875
9 18416.791015625
};
\addplot [semithick, pytorch]
table {%
3 1529
4 1539.00024414062
5 1583.00036621094
6 1798.99975585938
7 2667
8 6472.998046875
9 21519.01171875
};
\addplot [semithick, torchscript]
table {%
3 1356.99987792969
4 1356.99987792969
5 1359.00012207031
6 1379
7 1460.80187988281
8 1854.48486328125
9 3458.5263671875
10 9889.10546875
};
\end{axis}

\end{tikzpicture}
        \end{adjustbox}
    \end{subfigure}
    \begin{subfigure}{0.45\textwidth}
        \centering
        \begin{adjustbox}{width=\textwidth}
\begin{tikzpicture}

\begin{axis}[
axis background/.style={fill=white!89.8039215686275!black},
axis line style={white},
log basis y={10},
tick align=outside,
tick pos=left,
title={Average used memory for resolution $=64$},
x grid style={white},
xlabel={batch size},
xmajorgrids,
xmin=2.7, xmax=9.3,
xtick style={color=white!33.3333333333333!black},
y grid style={white},
ymajorgrids,
ymin=1000, ymax=4631.40123788891,
ymode=log,
ytick style={color=white!33.3333333333333!black},
xticklabels={$2^3$, $2^4$, $2^5$, $2^6$, $2^7$, $2^8$, $2^9$},
xtick={3,...,9},
]
\path [fill=cudnn, fill opacity=0.2, very thin]
(axis cs:3,1079)
--(axis cs:3,1079)
--(axis cs:4,1163)
--(axis cs:5,1291)
--(axis cs:6,1519)
--(axis cs:7,1961)
--(axis cs:8,2863)
--(axis cs:9,4321)
--(axis cs:9,4321)
--(axis cs:9,4321)
--(axis cs:8,2863)
--(axis cs:7,1961)
--(axis cs:6,1519)
--(axis cs:5,1291)
--(axis cs:4,1163)
--(axis cs:3,1079)
--cycle;

\path [fill=libtorch, fill opacity=0.2, very thin]
(axis cs:3,1551)
--(axis cs:3,1551)
--(axis cs:4,1573)
--(axis cs:5,1613)
--(axis cs:6,1669)
--(axis cs:7,1793)
--(axis cs:8,2105)
--(axis cs:9,2781)
--(axis cs:9,2781)
--(axis cs:9,2781)
--(axis cs:8,2105)
--(axis cs:7,1793)
--(axis cs:6,1669)
--(axis cs:5,1613)
--(axis cs:4,1573)
--(axis cs:3,1551)
--cycle;

\path [fill=pytorch, fill opacity=0.2, very thin]
(axis cs:3,1537)
--(axis cs:3,1537)
--(axis cs:4,1553)
--(axis cs:5,1583)
--(axis cs:6,1655)
--(axis cs:7,1801)
--(axis cs:8,2065)
--(axis cs:9,2621)
--(axis cs:9,2623)
--(axis cs:9,2623)
--(axis cs:8,2065)
--(axis cs:7,1801)
--(axis cs:6,1655)
--(axis cs:5,1583)
--(axis cs:4,1555)
--(axis cs:3,1537)
--cycle;

\path [fill=white!46.6666666666667!black, fill opacity=0.2, very thin]
(axis cs:3,1341)
--(axis cs:3,1341)
--(axis cs:4,1357)
--(axis cs:5,1359)
--(axis cs:6,1403)
--(axis cs:7,1391)
--(axis cs:8,1459.14)
--(axis cs:9,1633)
--(axis cs:9,1919)
--(axis cs:9,1919)
--(axis cs:8,1461)
--(axis cs:7,1391)
--(axis cs:6,1403)
--(axis cs:5,1359)
--(axis cs:4,1357)
--(axis cs:3,1341)
--cycle;

\path [fill=torchscript, fill opacity=0.2, very thin]
(axis cs:3,1341)
--(axis cs:3,1341)
--(axis cs:4,1357)
--(axis cs:5,1359)
--(axis cs:6,1403)
--(axis cs:7,1391)
--(axis cs:8,1459.14)
--(axis cs:9,1633)
--(axis cs:9,1919)
--(axis cs:9,1919)
--(axis cs:8,1461)
--(axis cs:7,1391)
--(axis cs:6,1403)
--(axis cs:5,1359)
--(axis cs:4,1357)
--(axis cs:3,1341)
--cycle;

\addplot [semithick, cudnn]
table {%
3 1079
4 1162.99987792969
5 1290.99975585938
6 1518.99963378906
7 1961.00012207031
8 2863.00073242188
9 4321.0009765625
};
\addplot [semithick, libtorch]
table {%
3 1551.00012207031
4 1573
5 1613.00024414062
6 1669.00048828125
7 1793.00012207031
8 2105.00048828125
9 2781
};
\addplot [semithick, pytorch]
table {%
3 1537
4 1554.19982910156
5 1583.00036621094
6 1654.99975585938
7 1800.99975585938
8 2064.99951171875
9 2621.69946289062
};
\addplot [semithick, torchscript]
table {%
3 1341
4 1356.99987792969
5 1359.00012207031
6 1403.00036621094
7 1391.00036621094
8 1460.81384277344
9 1886.31420898438
};
\end{axis}

\end{tikzpicture}
        \end{adjustbox}
    \end{subfigure}
    \caption{Comparison of execution time and memory efficiency for PyTorch, LibTorch, and cuDNN implementations on PASCAL for fixed batch size (32) and various resolutions and fixed resolution (64) and various batch sizes.}
    \label{fig:timing_results}
\end{figure*}
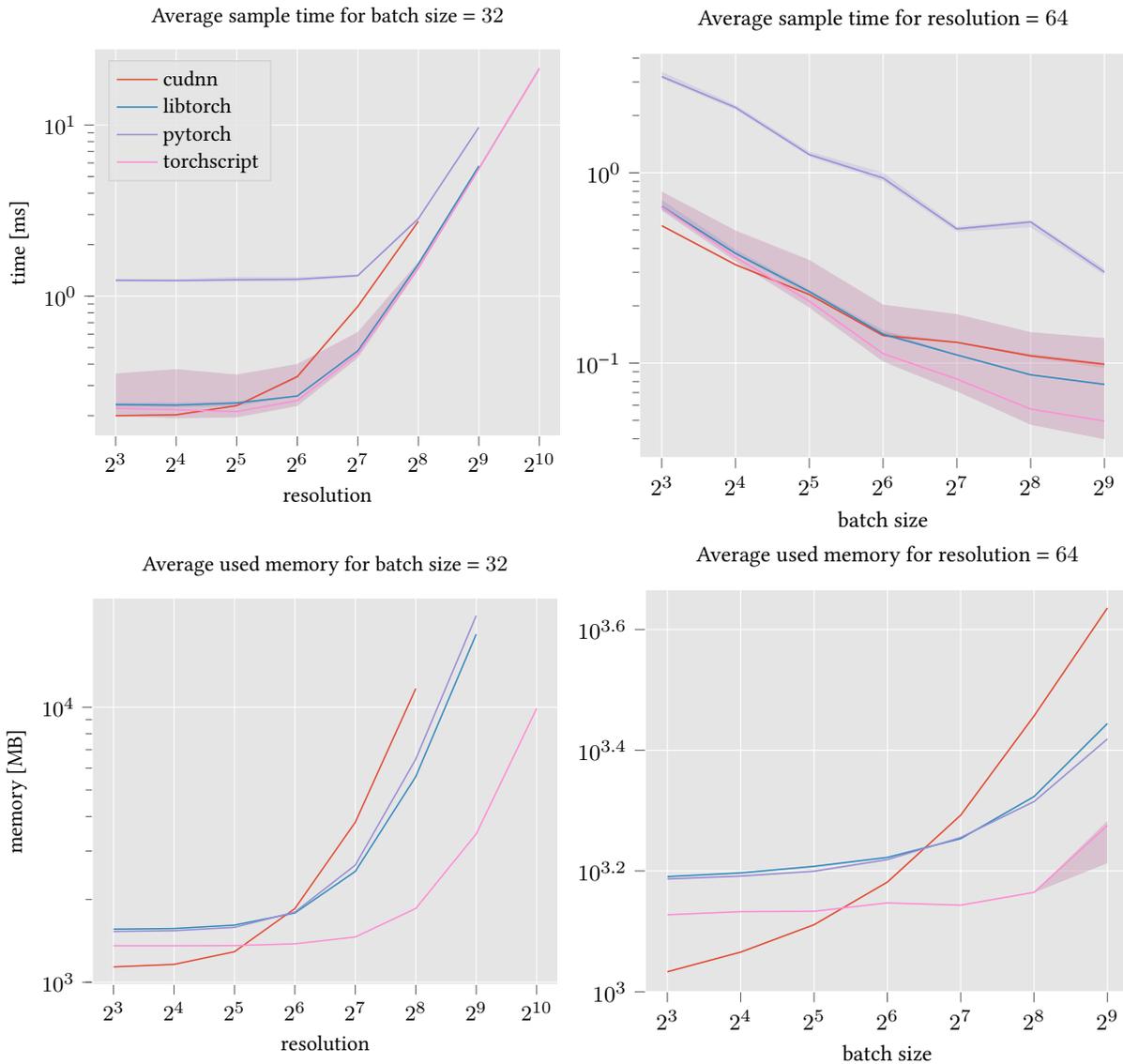

In terms of execution time and memory usage PyTorch compares unfavorably with each of the other implementations.
We measure execution time, memory usage, and GPU utilization during evaluation on PASCAL for various batch sizes and resolution.
For example, for fixed batch size and various resolutions and for fixed resolution and various batch sizes (see~\cref{fig:timing_results}), we see that PyTorch is almost an order of magnitude slower than all other implementations.
This execution time difference persists across resolutions and batch sizes but narrows as either increases (see~\cref{fig:train_avg_sample_time,,fig:train_avg_used_mem,,fig:train_avg_gpu_util,,fig:eval_avg_sample_time,,fig:eval_avg_used_mem,,fig:eval_avg_gpu_util} in the appendix).
With respect to memory usage PyTorch and LibTorch are approximately the same across resolutions and batch sizes, while cuDNN and TorchScript are more memory efficient, especially below resolution $2^6$ and batch size $2^7$.

We use NVIDIA's Visual profiler\footnote{\link{\url{https://developer.nvidia.com/nvidia-visual-profiler}}} to investigate fixed \code{batch\_size = 32} further.
One critical way in which the PyTorch implementation differs from the others is in host-to-device per batch copy size: PyTorch copies 25.166MB while the other implementations copy 12.583MB\@.
Consequently PyTorch requires $\sim$15.17ms to copy the batch while all other implementations require $\sim$7.55ms\footnote{In fact, pinning memory (copying from memory that is not paged) halves copy time again.}.
Another significant discrepancy is the choice in block size and grid size (regarding threaded distribution across GPU SMs).
For the ReLU kernel, which is the second most often executed kernel ($\sim$12\% of total execution time), the PyTorch implementation allocates a grid of size $\left( 1024,2,1 \right)$ while the LibTorch and cuDNN implementations allocate grids of size $\left( 512,2,1 \right)$.
Consequently, on average, each PyTorch ReLU invocation consumes $\sim$690.6 microseconds while each invocation consumes $\sim$372.6 microseconds.
it is unclear exactly why the larger grid leads to a slowdown but one hypothesis is that distributing work across more SMs leads to more stalls on cache misses\footnote{SM level statistics are not presented in the NVIDIA profiler.}.


\section{Discussion}\label{sec:discussion}

Overall the central challenges for this work revolved around the cuDNN implementation.
Chief among them involved familiarizing ourselves with the SIMT compute model and the CUDA/cuDNN APIs.
It is important to contextualize these challenges appropriately: how much of the challenge is akin to the initial learning curve associated with any new software toolset (e.g.\ PyTorch) and how much is enduring.
Certainly memory leaks, and debugging them, given that C++ is not a memory managed language, will persist, but the difficulties associated with the idiosyncrasies of the APIs most likely will not.
Assuming that the majority of challenge decreases with time, are these lower levels of abstraction worth the investment of effort and time?

The lackluster accuracy results of cuDNN seemingly do not bode well for the hypothesis that one can, in a straightforward way, trade performance for implementation effort.
The cuDNN performance indicates there are serious bugs with the implementation.
Alternatively the accuracy results suggest that there are optimizations present in the PyTorch and LibTorch implementations that are obscured from the user (such as the Kaiming initialization mentioned in~\cref{sec:results}).
The former case, when juxtaposed with the execution time and memory usage results, suggests that the cuDNN implementation could be as accurate the PyTorch and LibTorch implementations, with much lower execution time and memory usage (assuming the bugs can be rectified \textemdash\ a reasonable assumption).
In the latter case, we face a sort of existential crisis: how many results in DL research, attributed to architectural innovations, in fact, hinge on the implementation details of the frameworks those architectures are themselves implemented against?

It bears repetition: even broadly useful heuristics are a cost of abstraction if they cannot be adjusted.
Case in point, \textbf{Kaiming initialization is not always net positive with respect to performance}:
\begin{displayquote}[\cite{zhang2019fixup}]
    Standard initialization methods (Glorot \& Bengio, 2010; He et al., 2015; Xiao et al., 2018) attempt to  set  the  initial  parameters  of  the  network  such  that  the  activations  neither  vanish  nor  explode.
    Unfortunately, it has been observed that without normalization techniques such as BatchNorm they do not account properly for the effect of residual connections and this causes exploding gradients.
\end{displayquote}
In addition batch normalization layers being initialized to $(\gamma=1,\beta=0)$ also does not uniformly improve performance:
\begin{displayquote}[\cite{goyal2018accurate}]
    For BN layers, the learnable scaling coefficient $\gamma$ is initialized  to  be  1, \textit{except  for  each  residual  block’s  last  BN where $\gamma$ is initialized to be 0}.
    Setting $\gamma = 0$ in the last BN of each residual block causes the forward/backward signal initially to propagate through the identity shortcut of ResNets, which we found to ease optimization at the start of training.
\end{displayquote}

In theory, a sufficiently flexible DL compiler could rescue us from the purgatory we find ourselves in;
a sufficiently powerful compiler would implement the necessary DL abstractions in a robust way but also have enough flexibility to enable users to implement custom extensions of those abstractions.
One promising project that has as its goal such a high-level compiler is the ``Extensible Programming''~\cite{Besard_2019} project.
Besard et al. expose interfaces to alter the compilation process for Julia-lang\footnote{\link{\url{https://julialang.org/}}}.
The project instruments the Julia compiler itself and enables users to build high-level hardware abstractions in the source language itself%
\footnote{This is possible because Julia is JITed using LLVM and is homo-iconic i.e. it supports a macro system that can manipulate the JITed LLVM bytecode.}.
They've had initial success writing high-level GPU code that performs comparably with CUDA C\footnote{\link{\url{https://github.com/JuliaGPU/CUDAnative.jl}}}.

\section{Conclusion and future work}\label{sec:futurework}
In this work we have implemented ResNet-50 in PyTorch, LibTorch, TorchScript, and cuDNN.
We then trained\footnote{With the exception of TorchScript since currently it cannot be trained.} and evaluated each implementation on the MNIST, CIFAR10, STL10, and PASCAL VOC datasets.
Despite difficulties with the cuDNN implementation, we show that PyTorch underperforms lower level abstractions along various batch sizes and resolutions (see~\cref{fig:train_avg_sample_time,,fig:train_avg_used_mem,,fig:train_avg_gpu_util,,fig:eval_avg_sample_time,,fig:eval_avg_used_mem,,fig:eval_avg_gpu_util} in the appendix).
The ultimate causes for these differences in performance are hypothesized to be the larger buffers and larger grid allocations used by PyTorch and consequently longer host-to-device copy times.

Future work will focus on further narrowing down the causes of the memory and sample time inefficiencies of PyTorch;
in particular we hope to more closely investigate the execution paths of the PyTorch and LibTorch implementations in order to discover what additional heuristic choices are made (relative to the cuDNN implementation).
A long term research goal is to design a DL framework that uses code generation to statically generate C++ code corresponding to neural network graphs.
Such a framework would obviate the need for dynamic dispatch at the object level.

\section{Speculation}\label{sec:speculation}
We speculate about DL systems along three dimensions: hardware, software, and use cases/techniques.
Firstly, we project that with the end of Dennard scaling~\cite{CARDOSO201717} general purpose processors will give way to Application Specific Integrated Circuit (ASIC).
Architectures like Cerebras' CS-1, SambdaNova's Cardinal SN10, Google's TPU, and even Apple's Neural Engine (packaged with their M1) demonstrate that chip designers recognize the need for ML/DL purpose built hardware.
These purpose built chips are better suited for the concurrency and memory access patterns unique to ML/DL workloads.
With the impending cambrian explosion of different architectures, there will be a great need for compilers that act as intermediaries between high-level neural network representations and their hardware implementations.
Compiler infrastructures such as MLIR, PlaidML, and Intel's oneAPI~\cite{oneapi} will become critical for software developers.
As ML/DL ASICs become more ubiquitous and more performant, ML/DL powered software will become commensurately more ubiquitous;
already mobile phones employ ML/DL for text autocorrection~\cite{DBLP:journals/corr/abs-1709-06429}, image compression/super-resolution~\cite{DBLP:journals/corr/RomanoIM16}, and facial fingerprinting~\cite{cuda_toolkit}.
We project that most user-interaction driven tasks (e.g.\ setting alarms, coordinating appointments/meetings, planning daily routines) will have ML/DL solutions in the coming years.
Further, more platforms that have been up until today analog or simple computers (e.g.\ kitchen appliances, light power tools) will become ``smart''.
This distribution of lower power edge devices will necessitate more effective (and private) federated learning methods~\cite{48690}.

\section{Acknowledgements}\label{sec:acks}
We would like to thank Rick Stevens and Ian Foster for their constructive criticism and feedback on the project and paper itself.

\bibliographystyle{science}
\bibliography{main}

\begin{thebibliography}{10}

\bibitem{paszke2019pytorch}
Paszke, A. et~al, Pytorch: An imperative style, high-performance deep learning
  library (2019).

\bibitem{abadi2016tensorflow}
Abadi, M. et~al, Tensorflow: Large-scale machine learning on heterogeneous
  distributed systems (2016).

\bibitem{chen2015mxnet}
Chen, T. et~al, Mxnet: A flexible and efficient machine learning library for
  heterogeneous distributed systems (2015).

\bibitem{cntk}
Seide, F. and Agarwal, A., {\it Proceedings of the 22nd ACM SIGKDD
  International Conference on Knowledge Discovery and Data Mining\/}, KDD '16
  (Association for Computing Machinery, New York, NY, USA, 2016), p. 2135.

\bibitem{knuth}
Knuth, D.E., {\it Commun. ACM\/} {\bf 17}, 667–673 (1974).

\bibitem{abdelhamed2020ntire}
Abdelhamed, A. et~al, Ntire 2020 challenge on real image denoising: Dataset,
  methods and results (2020).

\bibitem{hall2020probability}
Hall, D. et~al, {\it IEEE Winter Conference on Applications of Computer Vision
  (WACV)\/} (2020).

\bibitem{ILSVRC15}
Russakovsky, O. et~al, {\it International Journal of Computer Vision (IJCV)\/}
  {\bf 115}, 211 (2015).

\bibitem{7780576}
{Shi}, W. et~al, {\it 2016 IEEE Conference on Computer Vision and Pattern
  Recognition (CVPR)\/} (2016), pp. 1874--1883.

\bibitem{brown2020language}
Brown, T.B. et~al, Language models are few-shot learners (2020).

\bibitem{Locke1689-LOCAEC-4}
Locke, J., {\it An Essay Concerning Human Understanding\/} (Oxford University
  Press, 1689).

\bibitem{Russell1937-RUSPOM-7}
Russell, B., {\it Principles of Mathematics\/} (Routledge, 1937).

\bibitem{abstraction}
Colburn, T. and Shute, G., {\it Minds and Machines\/} {\bf 17}, 169 (2007).

\bibitem{abelson1996structure}
Abelson, H., Sussman, G. and Sussman, J., {\it Structure and Interpretation of
  Computer Programs\/}, Electrical engineering and computer science series (MIT
  Press, 1996).

\bibitem{10.1145/244795.244798}
Pippenger, N., {\it ACM Trans. Program. Lang. Syst.\/} {\bf 19}, 223–238
  (1997).

\bibitem{10.5555/1410219}
Skiena, S.S., {\it The Algorithm Design Manual\/} (Springer Publishing Company,
  Incorporated, 2008), second edn.

\bibitem{10.1007/978-3-642-28869-2_1}
Stroustrup, B., {\it Proceedings of the 21st European Conference on Programming
  Languages and Systems\/}, ESOP'12 (Springer-Verlag, Berlin, Heidelberg,
  2012), p. 1–25.

\bibitem{10.5555/1891996}
Sanders, J. and Kandrot, E., {\it CUDA by Example: An Introduction to
  General-Purpose GPU Programming\/} (Addison-Wesley Professional, 2010), first
  edn.

\bibitem{5009071}
{Flynn}, M.J., {\it IEEE Transactions on Computers\/} {\bf C-21}, 948 (1972).

\bibitem{cuda_toolkit}
{NVIDIA}, Cuda toolkit documentation,
  \url{https://docs.nvidia.com/cuda/cuda-c-programming-guide/index.html\#control-flow-instructions}
  (2020). [Online; accessed 3-December-2020].

\bibitem{Glaskowsky2009NVIDIAS}
Glaskowsky, P. (2009).

\bibitem{5751939}
{Wittenbrink}, C.M., {Kilgariff}, E. and {Prabhu}, A., {\it IEEE Micro\/} {\bf
  31}, 50 (2011).

\bibitem{10.5555/3314872.3314876}
Panchenko, M. et~al, {\it Proceedings of the 2019 IEEE/ACM International
  Symposium on Code Generation and Optimization\/}, CGO 2019 (IEEE Press,
  2019), p. 2–14.

\bibitem{le2019tflms}
Le, T.D. et~al, Tflms: Large model support in tensorflow by graph rewriting
  (2019).

\bibitem{Pradelle2017PolyhedralOO}
Pradelle, B. et~al, {\it ESPT/VPA@SC\/} (2017).

\bibitem{lattner2020mlir}
Lattner, C. et~al, Mlir: A compiler infrastructure for the end of moore's law
  (2020).

\bibitem{zerrell2019stripe}
Zerrell, T. and Bruestle, J., Stripe: Tensor compilation via the nested
  polyhedral model (2019).

\bibitem{Griebl98codegeneration}
Griebl, M., Lengauer, C. and Wetzel, S., {\it In IEEE PACT\/} (IEEE Computer
  Society Press, 1998), pp. 106--111.

\bibitem{vasilache2018tensor}
Vasilache, N. et~al, Tensor comprehensions: Framework-agnostic high-performance
  machine learning abstractions (2018).

\bibitem{10.5555/3291168.3291211}
Chen, T. et~al, {\it Proceedings of the 13th USENIX Conference on Operating
  Systems Design and Implementation\/}, OSDI'18 (USENIX Association, USA,
  2018), p. 579–594.

\bibitem{he2015deep}
He, K. et~al, Deep residual learning for image recognition (2015).

\bibitem{he2015delving}
He, K. et~al, Delving deep into rectifiers: Surpassing human-level performance
  on imagenet classification (2015).

\bibitem{zhang2019fixup}
Zhang, H., Dauphin, Y.N. and Ma, T., Fixup initialization: Residual learning
  without normalization (2019).

\bibitem{goyal2018accurate}
Goyal, P. et~al, Accurate, large minibatch sgd: Training imagenet in 1 hour
  (2018).

\bibitem{Besard_2019}
Besard, T., Foket, C. and De~Sutter, B., {\it IEEE Transactions on Parallel and
  Distributed Systems\/} {\bf 30}, 827–841 (2019).

\bibitem{CARDOSO201717}
{\~a}~o M.P.~Cardoso, J., {\'e}~Gabriel F.~Coutinho, J. and Diniz, P.C., {\it
  Embedded Computing for High Performance\/}, {\~a}~o M.P.~Cardoso, J.,
  {\'e}~Gabriel F.~Coutinho, J. and Diniz, P.C., eds. (Morgan Kaufmann, Boston,
  2017), pp. 17 -- 56.

\bibitem{oneapi}
{Intel}, Driving a new era of accelerated computing,
  \url{https://software.intel.com/content/www/us/en/develop/tools/oneapi.html}
  (2020). [Online; accessed 3-December-2020].

\bibitem{DBLP:journals/corr/abs-1709-06429}
Ghosh, S. and Kristensson, P.O., {\it CoRR\/} {\bf abs/1709.06429} (2017).

\bibitem{DBLP:journals/corr/RomanoIM16}
Romano, Y., Isidoro, J. and Milanfar, P., {\it CoRR\/} {\bf abs/1606.01299}
  (2016).

\bibitem{48690}
Augenstein, S. et~al (2019).

\end{thebibliography}

\appendix
\appendixpage

\section{Extra plots}\label{sec:appendix}
\begin{figure*}
  \centering
  \begin{adjustbox}{height=.45\textheight}
    \begin{tikzpicture}[
    axisbg/.style={
        fill=#1!50,
        nearly transparent
    },
    declare function={
      ticklen=0.15;
      xmax=5;
      ymax=5;
      zmax=5;
    },
]

    \begin{axis}[%
        axis background/.style={fill=white!89.8039215686275!black},
        title={},
        grid=major,
        width=12cm,height=12cm,
        xlabel={batch size},
        ylabel={resolution},
        zlabel={util (\%)},
        legend style={
            mark size=5,
            legend cell align=left
        },
        legend entries={
            cuDNN, LibTorch, PyTorch, TorchScript
        },
        label style={font=\scriptsize},
        ticklabel style={font=\scriptsize},
        view={40}{30},
        xmode=log,
        ymode=log,
        log basis x={2},
        log basis y={2},
        zmode=log,
        log origin=infty
    ]
        \addplot3 [
            opacity=0.7,
            ycomb,
            line width=0.5pt,
            mark=cube*,
            mark size=3,
            fill=cudnn
        ]
        file{./figures/plots/cudnn_train_avg_gpu_util.csv};

        \addplot3 [
            opacity=0.7,
            ycomb,
            line width=0.5pt,
            mark=ball,
            mark size=3,
            fill=libtorch,
        ]
        file{./figures/plots/libtorch_train_avg_gpu_util.csv};

        \addplot3 [
            opacity=0.7,
            ycomb,
            line width=0.5pt,
            mark=diamond*,
            mark size=3,
            fill=pytorch,
        ]
        file{./figures/plots/pytorch_train_avg_gpu_util.csv};

    \end{axis}
\end{tikzpicture}
  \end{adjustbox}
  \caption{Average GPU utilization during in training}\label{fig:train_avg_gpu_util}
\end{figure*}
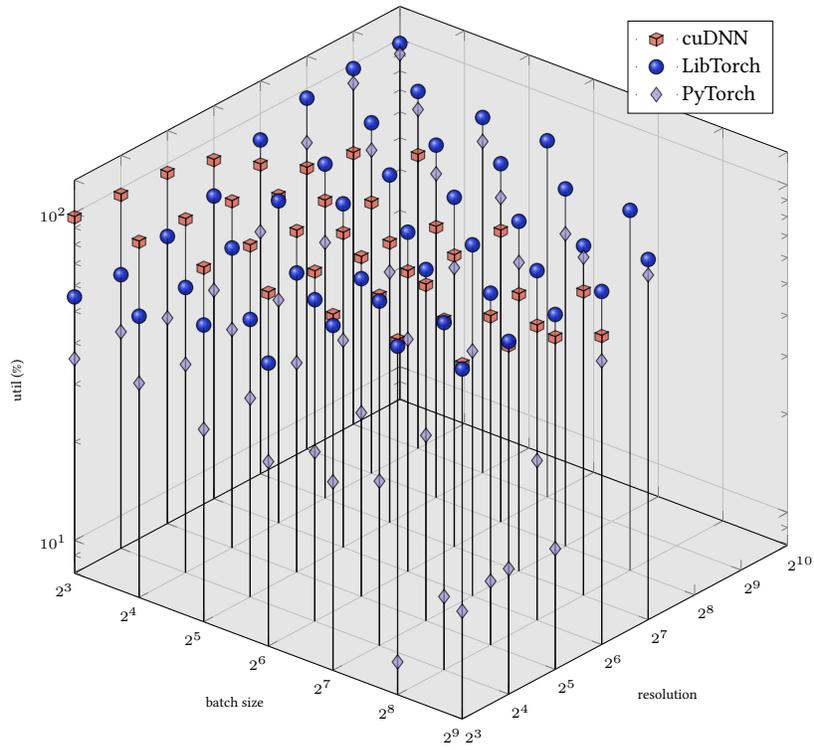

\bigskip

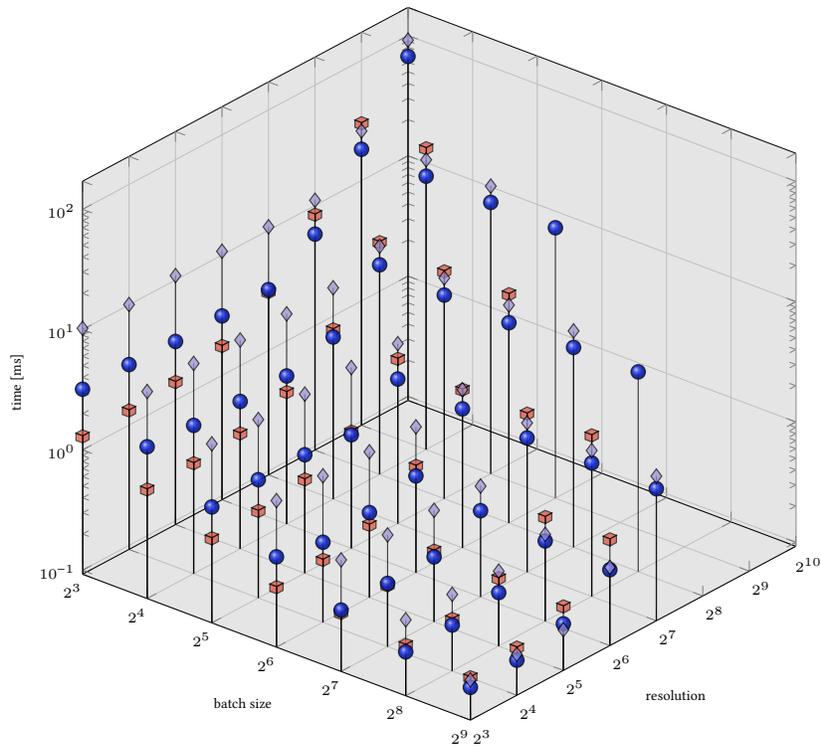
\begin{figure*}
  \centering
  \begin{adjustbox}{height=.45\textheight}
    \begin{tikzpicture}[
    axisbg/.style={
        fill=#1!50,
        nearly transparent
    },
    declare function={
      ticklen=0.15;
      xmax=5;
      ymax=5;
      zmax=5;
    },
]

    \begin{axis}[%
        axis background/.style={fill=white!89.8039215686275!black},
        title={},
        grid=major,
        width=12cm,height=12cm,
        xlabel={batch size},
        ylabel={resolution},
        zlabel={time [ms]},
        label style={font=\scriptsize},
        ticklabel style={font=\scriptsize},
        view={40}{30},
        xmode=log,
        ymode=log,
        log basis x={2},
        log basis y={2},
        zmode=log,
        log origin=infty
    ]
        \addplot3 [
            opacity=0.7,
            ycomb,
            line width=0.5pt,
            mark=cube*,
            mark size=3,
            fill=cudnn
        ]
        file{./figures/plots/cudnn_train_avg_sample_time.csv};

        \addplot3 [
            opacity=0.7,
            ycomb,
            line width=0.5pt,
            mark=ball,
            mark size=3,
            fill=libtorch,
        ]
        file{./figures/plots/libtorch_train_avg_sample_time.csv};

        \addplot3 [
            opacity=0.7,
            ycomb,
            line width=0.5pt,
            mark=diamond*,
            mark size=3,
            fill=pytorch,
        ]
        file{./figures/plots/pytorch_train_avg_sample_time.csv};

    \end{axis}
\end{tikzpicture}
  \end{adjustbox}
  \caption{Average sample time in training on PASCAL}\label{fig:train_avg_sample_time}
\end{figure*}

\bigskip

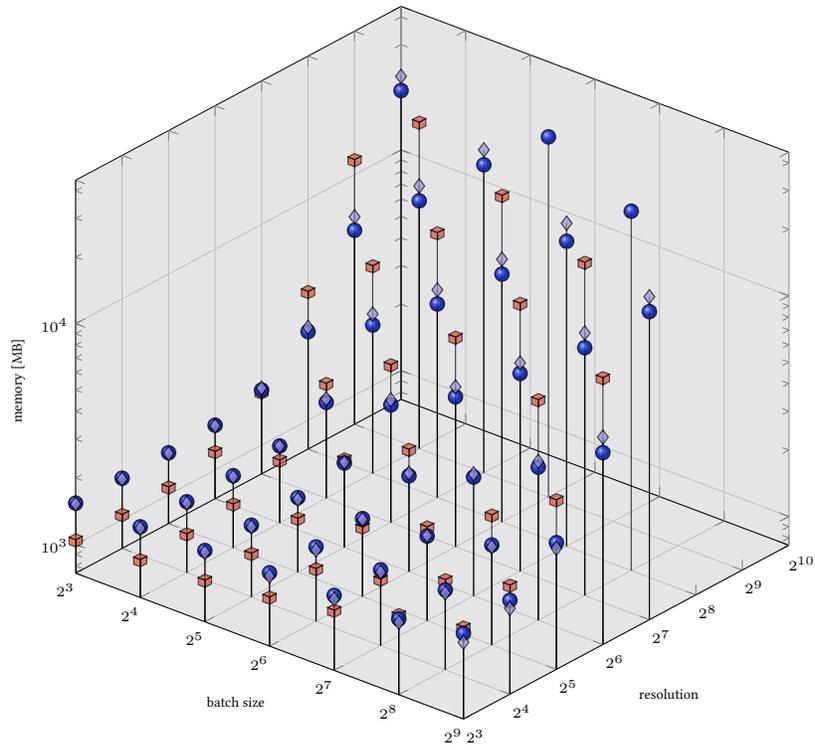
\begin{figure*}
  \centering
  \begin{adjustbox}{height=.45\textheight}
    \begin{tikzpicture}[
    axisbg/.style={
        fill=#1!50,
        nearly transparent
    },
    declare function={
      ticklen=0.15;
      xmax=5;
      ymax=5;
      zmax=5;
    },
]

    \begin{axis}[%
        axis background/.style={fill=white!89.8039215686275!black},
        title={},
        grid=major,
        width=12cm,height=12cm,
        xlabel={batch size},
        ylabel={resolution},
        zlabel={memory [MB]},
        label style={font=\scriptsize},
        ticklabel style={font=\scriptsize},
        view={40}{30},
        xmode=log,
        ymode=log,
        log basis x={2},
        log basis y={2},
        zmode=log,
        log origin=infty
    ]
        \addplot3 [
            opacity=0.7,
            ycomb,
            line width=0.5pt,
            mark=cube*,
            mark size=3,
            fill=cudnn
        ]
        file{./figures/plots/cudnn_train_avg_used_mem.csv};

        \addplot3 [
            opacity=0.7,
            ycomb,
            line width=0.5pt,
            mark=ball,
            mark size=3,
            fill=libtorch,
        ]
        file{./figures/plots/libtorch_train_avg_used_mem.csv};

        \addplot3 [
            opacity=0.7,
            ycomb,
            line width=0.5pt,
            mark=diamond*,
            mark size=3,
            fill=pytorch,
        ]
        file{./figures/plots/pytorch_train_avg_used_mem.csv};

    \end{axis}
\end{tikzpicture}
  \end{adjustbox}
  \caption{Average memory used in training on PASCAL}\label{fig:train_avg_used_mem}
\end{figure*}

\bigskip

\begin{figure*}
  \centering
  \begin{adjustbox}{height=.45\textheight}
    \begin{tikzpicture}[
    axisbg/.style={
        fill=#1!50,
        nearly transparent
    },
    declare function={
      ticklen=0.15;
      xmax=5;
      ymax=5;
      zmax=5;
    },
]

    \begin{axis}[%
        axis background/.style={fill=white!89.8039215686275!black},
        title={},
        grid=major,
        width=12cm,height=12cm,
        xlabel={batch size},
        ylabel={resolution},
        zlabel={util (\%)},
        legend style={
            mark size=5,
            legend cell align=left
        },
        legend entries={
            cuDNN, LibTorch, PyTorch, TorchScript
        },
        label style={font=\scriptsize},
        ticklabel style={font=\scriptsize},
        view={40}{30},
        xmode=log,
        ymode=log,
        log basis x={2},
        log basis y={2},
        zmode=log,
        log origin=infty
    ]
        \addplot3 [
            opacity=0.7,
            ycomb,
            line width=0.5pt,
            mark=cube*,
            mark size=3,
            fill=cudnn
        ]
        file{./figures/plots/cudnn_eval_avg_gpu_util.csv};

        \addplot3 [
            opacity=0.7,
            ycomb,
            line width=0.5pt,
            mark=ball,
            mark size=3,
            fill=libtorch,
        ]
        file{./figures/plots/libtorch_eval_avg_gpu_util.csv};

        \addplot3 [
            opacity=0.7,
            ycomb,
            line width=0.5pt,
            mark=diamond*,
            mark size=3,
            fill=pytorch,
        ]
        file{./figures/plots/pytorch_eval_avg_gpu_util.csv};

        \addplot3 [
            opacity=0.7,
            ycomb,
            line width=0.5pt,
            mark=triangle*,
            mark size=3,
            fill=torchscript,
        ]
        file{./figures/plots/torchscript_eval_avg_gpu_util.csv};
    \end{axis}
\end{tikzpicture}
  \end{adjustbox}
  \caption{Average GPU utilization during in evaluation on PASCAL}\label{fig:eval_avg_gpu_util}
\end{figure*}
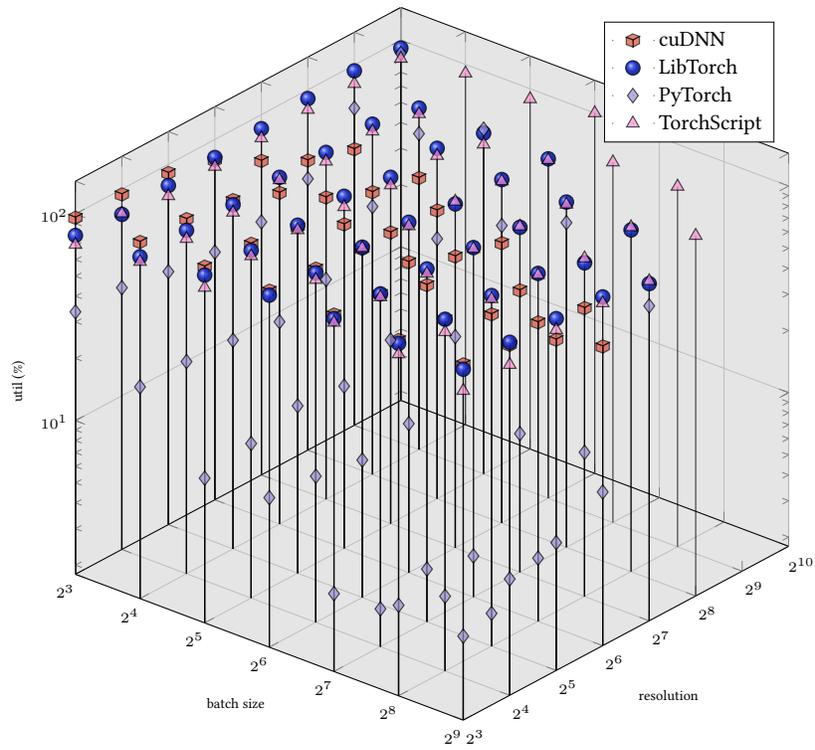

\bigskip

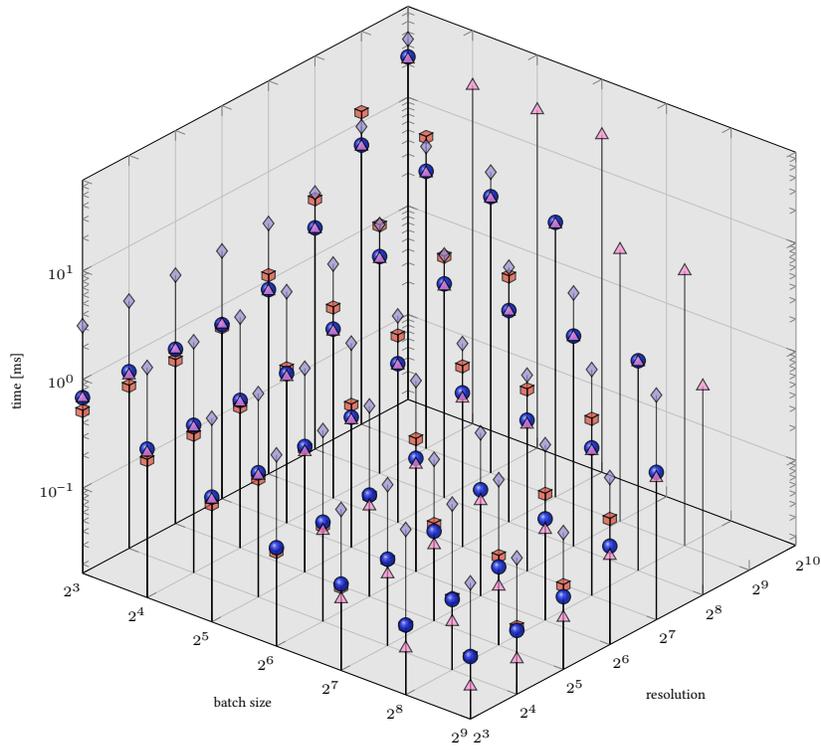
\begin{figure*}
  \centering
  \begin{adjustbox}{height=.45\textheight}
    \begin{tikzpicture}[
    axisbg/.style={
        fill=#1!50,
        nearly transparent
    },
    declare function={
      ticklen=0.15;
      xmax=5;
      ymax=5;
      zmax=5;
    },
]

    \begin{axis}[%
        axis background/.style={fill=white!89.8039215686275!black},
        title={},
        grid=major,
        width=12cm,height=12cm,
        xlabel={batch size},
        ylabel={resolution},
        zlabel={time [ms]},
        label style={font=\scriptsize},
        ticklabel style={font=\scriptsize},
        view={40}{30},
        xmode=log,
        ymode=log,
        log basis x={2},
        log basis y={2},
        zmode=log,
        log origin=infty
    ]
        \addplot3 [
            opacity=0.7,
            ycomb,
            line width=0.5pt,
            mark=cube*,
            mark size=3,
            fill=cudnn
        ]
        file{./figures/plots/cudnn_eval_avg_sample_time.csv};

        \addplot3 [
            opacity=0.7,
            ycomb,
            line width=0.5pt,
            mark=ball,
            mark size=3,
            fill=libtorch,
        ]
        file{./figures/plots/libtorch_eval_avg_sample_time.csv};

        \addplot3 [
            opacity=0.7,
            ycomb,
            line width=0.5pt,
            mark=diamond*,
            mark size=3,
            fill=pytorch,
        ]
        file{./figures/plots/pytorch_eval_avg_sample_time.csv};

        \addplot3 [
            opacity=0.7,
            ycomb,
            line width=0.5pt,
            mark=triangle*,
            mark size=3,
            fill=torchscript,
        ]
        file{./figures/plots/torchscript_eval_avg_sample_time.csv};
    \end{axis}
\end{tikzpicture}
  \end{adjustbox}
  \caption{Average sample time in evaluation on PASCAL}\label{fig:eval_avg_sample_time}
\end{figure*}

\bigskip

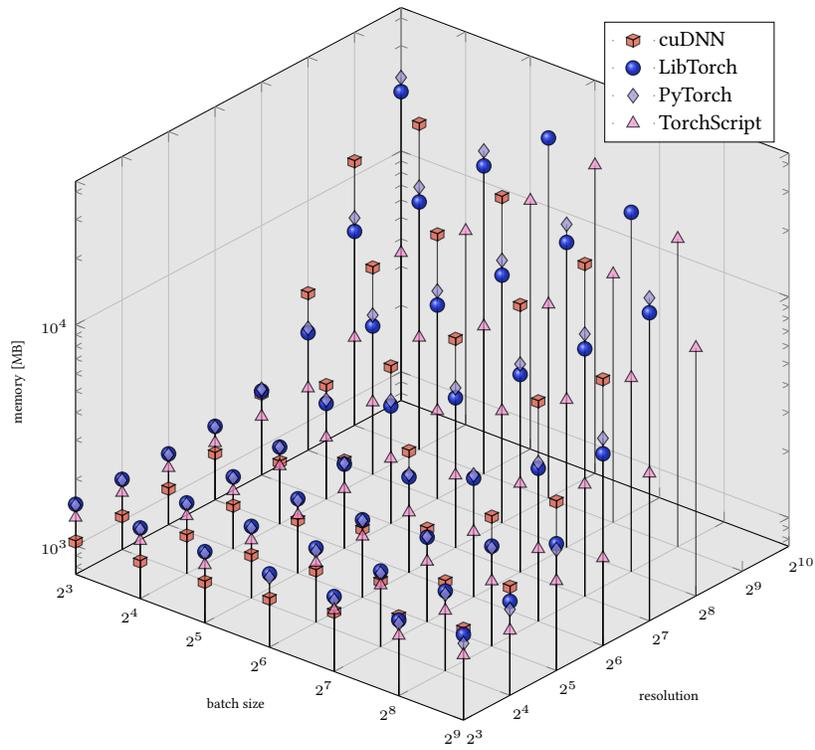
\begin{figure*}
  \centering
  \begin{adjustbox}{height=.45\textheight}
    \begin{tikzpicture}[
    axisbg/.style={
        fill=#1!50,
        nearly transparent
    },
    declare function={
      ticklen=0.15;
      xmax=5;
      ymax=5;
      zmax=5;
    },
]

    \begin{axis}[%
        axis background/.style={fill=white!89.8039215686275!black},
        title={},
        grid=major,
        width=12cm,height=12cm,
        xlabel={batch size},
        ylabel={resolution},
        zlabel={memory [MB]},
        legend style={
            mark size=5,
            legend cell align=left
        },
        legend entries={
            cuDNN, LibTorch, PyTorch, TorchScript
        },
        label style={font=\scriptsize},
        ticklabel style={font=\scriptsize},
        view={40}{30},
        xmode=log,
        ymode=log,
        log basis x={2},
        log basis y={2},
        zmode=log,
        log origin=infty
    ]
        \addplot3 [
            opacity=0.7,
            ycomb,
            line width=0.5pt,
            mark=cube*,
            mark size=3,
            fill=cudnn
        ]
        file{./figures/plots/cudnn_eval_avg_used_mem.csv};

        \addplot3 [
            opacity=0.7,
            ycomb,
            line width=0.5pt,
            mark=ball,
            mark size=3,
            fill=libtorch,
        ]
        file{./figures/plots/libtorch_eval_avg_used_mem.csv};

        \addplot3 [
            opacity=0.7,
            ycomb,
            line width=0.5pt,
            mark=diamond*,
            mark size=3,
            fill=pytorch,
        ]
        file{./figures/plots/pytorch_eval_avg_used_mem.csv};

        \addplot3 [
            opacity=0.7,
            ycomb,
            line width=0.5pt,
            mark=triangle*,
            mark size=3,
            fill=torchscript,
        ]
        file{./figures/plots/torchscript_eval_avg_used_mem.csv};
    \end{axis}
\end{tikzpicture}
  \end{adjustbox}
  \caption{Average memory used in evaluation on PASCAL}\label{fig:eval_avg_used_mem}
\end{figure*}

\begin{figure*}
    \centering
    \begin{subfigure}{0.45\linewidth}
        \centering
        \begin{adjustbox}{width=\textwidth}
            \input{figures/plots/train_accuracy_per_epoch_stl10.tex}
        \end{adjustbox}
    \end{subfigure}%
    \begin{subfigure}{0.45\textwidth}
        \centering
        \begin{adjustbox}{width=\textwidth}
            \input{figures/plots/train_accuracy_per_epoch_pascal.tex}
        \end{adjustbox}
    \end{subfigure}
    \quad
    \begin{subfigure}{0.45\textwidth}
        \centering
        \begin{adjustbox}{width=\textwidth}
            \input{figures/plots/eval_accuracy_per_epoch_stl10.tex}
        \end{adjustbox}
    \end{subfigure}
    \begin{subfigure}{0.45\textwidth}
        \centering
        \begin{adjustbox}{width=\textwidth}
            \input{figures/plots/eval_accuracy_per_epoch_pascal.tex}
        \end{adjustbox}
    \end{subfigure}
    \caption{Comparison of accuracy for PyTorch, LibTorch, and cuDNN implementations during training and evaluation. Solid line corresponds to mean while shaded regions correspond to min and max. Time is measured in units of \textit{epoch} $\times$ \textit{average epoch time}. }
    \label{fig:other_accuracy_results}
\end{figure*}

\end{document}